\documentclass{article}
\usepackage{mathptmx}
\usepackage{arxiv}
\usepackage{amsmath,amssymb}
\usepackage{booktabs,array,tabularx}
\usepackage{graphicx}
\usepackage{tikz}
\usepackage[numbers,sort&compress]{natbib}
\usepackage{enumitem}
\usepackage[expansion=false]{microtype}
\usepackage{textcomp}
\usepackage{caption}
\usepackage{xurl}
\usepackage[colorlinks=true,linkcolor=black,citecolor=black,urlcolor=black]{hyperref}
\definecolor{v1C}{HTML}{A85677}         % V1 Technical Failure (rose)
\definecolor{v2C}{HTML}{599E69}         % V2 Deployment Controls (green)
\definecolor{v3C}{HTML}{4980B0}         % V3 Consequence and Impact (steel blue)
\definecolor{tealC}{HTML}{3C898A}       % applicability strip
\definecolor{darkBlue}{HTML}{11235F}    % system-level evaluation bar
\definecolor{dgC}{HTML}{464646}         % outlines, equations, shared elements
\definecolor{evalindigo}{HTML}{534AB7}  % Build, Break, Eval, Fix
\definecolor{stSup}{HTML}{0F6E56}   % Supported
\definecolor{stRef}{HTML}{A63A32}   % Refuted, invalid
\definecolor{stUnr}{HTML}{9A6B00}   % Unresolved, review
\definecolor{stNA}{HTML}{5F5E5A}    % no evidence, not applicable, no effect

\usetikzlibrary{arrows.meta,positioning,fit,calc,backgrounds,shapes.geometric,decorations.pathreplacing}
\tikzset{
  >=Stealth,
  box/.style={draw=dgC!75, rounded corners=2pt, fill=white, align=center, font=\footnotesize,
              inner sep=3pt, minimum height=7mm, line width=0.5pt, text=dgC},
  proc/.style={box, draw=darkBlue, fill=darkBlue!6},          % shared evidence and decision processes
  dec/.style={box, draw=darkBlue, fill=darkBlue!14, line width=0.9pt},  % authorization dispositions
  v1box/.style={box, draw=v1C, fill=v1C!10},         % V1 technical failure
  v2box/.style={box, draw=v2C, fill=v2C!10},         % V2 deployment controls
  v3box/.style={box, draw=v3C, fill=v3C!10},                  % V3 consequence and impact
  vbox/.style={proc}, ctl/.style={v2box}, harm/.style={v3box},
  sSup/.style={box, draw=stSup, fill=stSup!12, line width=0.9pt, font=\footnotesize\bfseries},
  sRef/.style={box, draw=stRef, fill=stRef!12, line width=0.9pt, font=\footnotesize\bfseries, dash pattern=on 4pt off 1.5pt},
  sUnr/.style={box, draw=stUnr, fill=stUnr!14, line width=0.9pt, font=\footnotesize\bfseries, dash pattern=on 1.5pt off 1.5pt},
  sNA/.style={box, draw=stNA, fill=white, line width=0.7pt, font=\footnotesize, text=dgC!80},
  note/.style={font=\scriptsize, align=center, text=dgC!85},
  flow/.style={-{Stealth[length=2.2mm]}, line width=0.7pt},
  cflow/.style={-{Stealth[length=2.2mm]}, line width=1.4pt},
  dflow/.style={-{Stealth[length=2.2mm]}, line width=0.6pt, dash pattern=on 3pt off 2pt},
  dotflow/.style={-{Stealth[length=2.0mm]}, line width=0.6pt, dash pattern=on 1pt off 2pt},
  tagInv/.style={font=\scriptsize\bfseries, draw=stRef, fill=stRef!14, text=stRef!85!black, inner sep=2pt, rounded corners=1pt, line width=0.8pt},
  tagRev/.style={font=\scriptsize\bfseries, draw=stUnr, fill=stUnr!14, text=stUnr!85!black, inner sep=2pt, rounded corners=1pt, dash pattern=on 2pt off 1pt},
  tagNo/.style={font=\scriptsize\bfseries, draw=stNA, fill=white, text=stNA!90!black, inner sep=2pt, rounded corners=1pt},
  tagRet/.style={font=\scriptsize\bfseries, draw=darkBlue, fill=darkBlue!8, text=darkBlue, inner sep=2pt, rounded corners=1pt, dash pattern=on 1pt off 1pt},
  synth/.style={font=\scriptsize\bfseries, draw=black, fill=white, inner sep=2pt, rounded corners=1pt},
}

\setlist{nosep,leftmargin=1.4em}

\newcommand{\RBED}{R-BED}
\newcommand{\RBEDtm}{R-BED\texttrademark}
\newcommand{\Ex}{\mathrm{E}}
\newcommand{\Prob}{\mathrm{P}}
\newcommand{\cnd}{\,\vert\,}
\newcommand{\synth}{\textbf{SYNTHETIC}}
\newcolumntype{L}[1]{>{\raggedright\arraybackslash}p{#1}}

\title{R-BED\textsuperscript{TM}: A Consequence-Aware Framework for System-Level AI Evaluation and Continuous Assurance}
\author{Vishal Kumar Srivastava, PhD\\ Independent Researcher\\ \texttt{vshfrm@gmail.com}}
\renewcommand{\shorttitle}{R-BED: Consequence-Aware System-Level AI Evaluation}
\date{5 November 2026}

\begin{document}
\maketitle
\begin{abstract}
Evaluations of AI systems usually report properties of a model, while decisions to deploy concern a configured system: a model together with prompts, tools, permissions, data, safeguards and an operating context. We ask how evidence about technical failures, deployment-control effectiveness and real-world consequences can be connected to defensible authorization decisions throughout the lifecycle of a deployed system. We present \RBED{} (Risk-Based Evaluability Design), a specification-level methodology with a fixed 6+4+3 architecture of thirteen diagnostic dimensions (technical failure, deployment controls, consequence and impact); a failure-mediated expected-loss identity that is kept separate from intended-behavior and misuse pathways; bounded evidence statuses with independence levels; a control assurance ladder; five authorization dispositions governed by eight rules; and a Continuous System Evaluation (CSE) layer in which evidence carries a configuration pin, an expiry and declared dependencies, so that a change invalidates evidence in proportion rather than wholesale. We work through a synthetic refund-agent walkthrough, with a deterministic script that reproduces every number, to show how a task-success view can miss a consequential control weakness that the procedure identifies as missing evidence. The contribution is a reproducible methodological specification. Most components are standard practice integrated into one evidence-to-decision record; the Evidence Validity Register, the authority envelope and the tier rules are proposals. None has been empirically validated, and we state seven hypotheses for independent testing.
\end{abstract}
\vspace{2pt}

\section{Introduction}\label{sec:intro}
A deployment decision is a decision about a system, not about a model. A language model that scores well on a benchmark can still be configured with a tool that moves money, a retrieval source that carries untrusted text, an approval step that reviewers treat as a formality and a monitor that nobody owns. The consequences of such a system depend on what it is allowed to do, what stands between a failure and a person, and how long the evidence about it stays true. Evaluation practice, standards and regulatory guidance increasingly recognize this \citep{nist_rmf,nist_genai,weidinger2023}, but the methods that turn test results into a decision are usually left to the adopting organization.

This paper is organized around one question.
\begin{quote}\itshape How can evidence about technical AI failures, deployment-control effectiveness and real-world consequences be connected to defensible authorization decisions throughout the lifecycle of a deployed AI system?\end{quote}

\begin{figure}[!tbp]
\begingroup\hyphenpenalty=10000 \exhyphenpenalty=10000 \centering\resizebox{1.0\linewidth}{!}{\begin{tikzpicture}[node distance=7mm, every node/.append style={}]
  \tikzset{eq/.style={text width=26mm, minimum width=30mm, minimum height=21mm}}
  \node[v1box, eq] (m) {Model behavior\\ \scriptsize (V1 technical failure)};
  \node[box, right=8mm of m, eq] (c) {Deployment configuration\\ \scriptsize tools, permissions, prompts, data};
  \node[ctl, right=8mm of c, eq] (k) {Deployment controls\\ \scriptsize (V2: B1 to B4)};
  \node[harm, right=8mm of k, eq] (q) {Consequences\\ \scriptsize (V3: C1 to C3)};
  \node[dec, right=8mm of q, eq] (d) {Authorization decision\\ \scriptsize five dispositions};
  \draw[flow] (m)--(c); \draw[flow] (c)--(k); \draw[flow] (k)--(q); \draw[flow] (q)--(d);
  % intended behavior / misuse bypass (no technical failure)
  \draw[dflow] (c.south) -- ++(0,-9mm) -| node[pos=0.25, below, note, text width=44mm] {harm without technical failure:\\ intended behavior (N) or misuse (M)} (q.south);
  % scope brackets
  \draw[decorate, decoration={brace, amplitude=3pt}, line width=0.5pt] ([yshift=3mm]m.north west) -- node[above=2pt, note] {model-level metrics} ([yshift=3mm]m.north east);
  \draw[decorate, decoration={brace, amplitude=3pt}, line width=0.5pt] ([yshift=3mm]c.north west) -- node[above=2pt, note] {system-level evidence needed} ([yshift=3mm]d.north east);
  \node[note, below=2pt of d.south, text width=30mm] {records evidence, uncertainty and expiry};
\end{tikzpicture}}\par\endgroup
\caption{The evaluation-to-consequence gap. Model-level metrics cover the first stage. A defensible authorization needs evidence across the configuration, the controls that limit what a failure can do, and the consequences that remain. The dashed path shows that harm can arise without a technical failure. Dimension codes (V1 to V3, B1 to B4, C1 to C3) are defined in Section~\ref{sec:arch}.}
\label{fig:gap}
\end{figure}
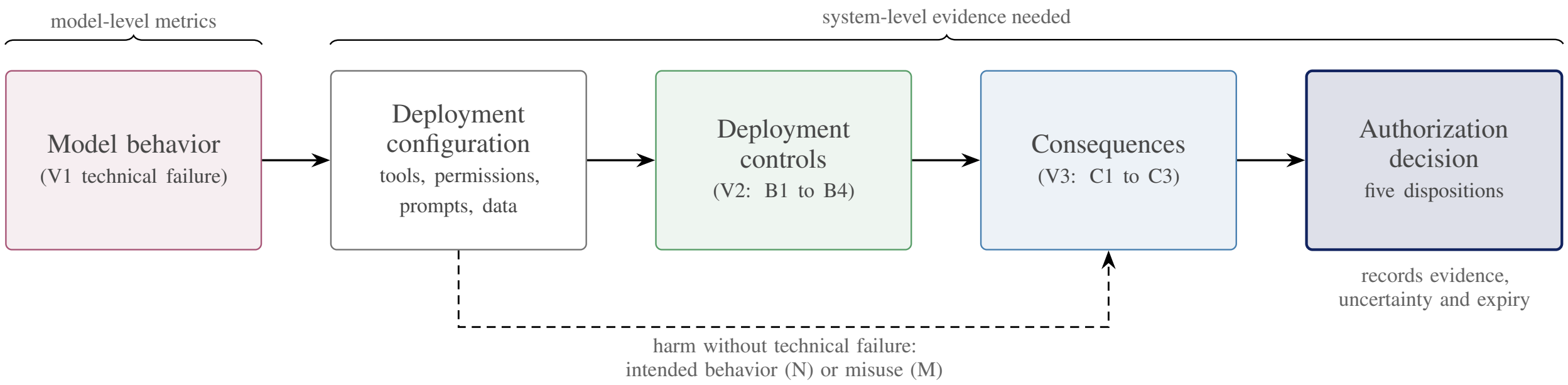

\paragraph{Three gaps.} We motivate the question with three gaps (Figure~\ref{fig:gap}).
\begin{itemize}
  \item \textbf{Evaluation gap.} Model-level metrics do not establish system-level performance or assurance. A model that behaves acceptably in isolation can violate its authority once tools, permissions and untrusted inputs are attached.
  \item \textbf{Consequence gap.} A failure and a harm are different events. A failure can be contained before it harms anyone, and harm can occur without any technical failure, through intended behavior or misuse.
  \item \textbf{Evidence-to-decision gap.} Testing and monitoring do not by themselves determine whether a system should be deployed, restricted, remediated or stopped, and the evidence behind a decision ages and is invalidated by change.
\end{itemize}

\paragraph{Approach.} \RBEDtm{} addresses the gaps with a single traceable record of a configured system: its claims, the evidence for each claim with a status, an independence level and an expiry, the controls on which the risk argument relies and the level to which each is shown to work, the consequences at stake, and the accountable decision. It builds on a conditional expected-loss decomposition of automation-related loss into failure, propagation to harm and severity \citep{srivastava2026}, and it connects that decomposition to evidence, controls and authorization rules. The methodology is published as three companion specifications, a public framework, a Continuous System Evaluation (CSE) specification and a crosswalk to the NIST AI RMF \citep{rbed_framework,rbed_cse,rbed_crosswalk}.

\paragraph{What is and is not claimed.} We do not claim that \RBED{} improves decisions or outcomes, and we do not claim novelty for its ingredients: risk as likelihood and consequence \citep{kaplan1981}, layered protection, binomial bounds \citep{clopper1934}, model validation and monitoring \citep{sr117}, assurance cases \citep{clymer2024} and continuous assurance all have precedent. The proposed contribution is their integration in one evidence-to-decision record for configured, including agentic, systems, with three elements we believe are specific to this specification and untested: evidence records with proportional, dependency-based invalidation; an authority envelope with trajectory metrics; and ordinal monitoring-tier rules (Section~\ref{sec:cse}). Components carry a maturity label: \textbf{[E]} specified core, with software checks of the specified rules in the author's reference tooling; \textbf{[P]} proposed extension; \textbf{[H]} research hypothesis requiring future testing. A label describes the state of specification and testing, not validated effectiveness.

\paragraph{Contributions.} The contributions are methodological and specification-level.
\begin{enumerate}
  \item A fixed architecture of thirteen diagnostic dimensions in three vertices (6+4+3) and five cross-cutting objectives, designed so that evidence from one vertex cannot substitute for evidence from another and no score is formed (Section~\ref{sec:arch}).
  \item A formal treatment of failure-mediated expected loss as a conditional identity, kept separate from intended-behavior and misuse pathways, with bounds for dependence, uncertainty and overlap, and an explicit distinction between expected, stressed and tail loss (Section~\ref{sec:risk}).
  \item A separation of test result, evidence assessment and deployment decision, with four bounded evidence statuses, independence levels I0 to I3, a control assurance ladder L0 to L3 distinct from independence, five dispositions and eight authorization rules (Section~\ref{sec:evidence}).
  \item A Continuous System Evaluation layer that makes evidence temporal: configuration-pinned evidence with review dates, expiry and declared dependencies, proportional invalidation, independent control challenge, agentic authority-envelope evaluation, consequence-prioritized monitoring tiers and a gated failure-to-learning loop (Section~\ref{sec:cse}).
  \item A reproducible, clearly synthetic walkthrough with a deterministic script, and a validation agenda of seven hypotheses (Sections~\ref{sec:case} and~\ref{sec:discussion}).
\end{enumerate}

\section{Background and Related Work}\label{sec:related}
We summarize the closest bodies of work and then position \RBED{} against them. The descriptions are the author's summaries of public sources; readers should consult each source for its authoritative scope.

\paragraph{Frameworks and standards.} The NIST AI Risk Management Framework organizes voluntary outcomes under Govern, Map, Measure and Manage and treats post-deployment monitoring as an outcome \citep{nist_rmf}; its Generative AI Profile adds suggested actions for generative systems \citep{nist_genai}; ISO/IEC 42001 specifies a certifiable AI management system \citep{iso42001}. These define what an organization should achieve and govern, not the evidence procedure that connects a test to an authorization. \RBED{} is written to supply system-level evidence for several outcomes; a companion crosswalk maps 51 of the 72 AI RMF subcategories (relevant or partial), without claiming coverage or conformity \citep{rbed_crosswalk}.

\paragraph{Evaluation, assurance cases and model risk.} Holistic benchmarks evaluate language models across scenarios \citep{liang2023helm}; extreme-risk evaluation argues that capability and alignment evaluations should inform responsible decisions \citep{shevlane2023}; a sociotechnical view distinguishes capability, human-interaction and systemic-impact layers \citep{weidinger2023}. A safety case is a structured argument that a system is acceptably safe to deploy \citep{clymer2024}; \RBED{} fixes record fields, statuses and rules so that the argument is checkable for discipline, though a record that satisfies the rules is not thereby safe. Supervisory guidance on model risk management requires validation before use, ongoing monitoring, outcomes analysis and documented change decisions \citep{sr117}; interagency guidance of April 2026 replaced SR 11-7 and, in its footnote 3, places generative and agentic AI models outside its scope \citep{sr262}.

\paragraph{Agentic evaluation, monitoring and learning from failure.} Agent benchmarks evaluate policy-following and reliability \citep{yao2024tau}, prompt-injection attacks and defenses \citep{debenedetti2024} and risks in emulated tool sandboxes \citep{ruan2024}. These supply test constructions; \RBED{} specifies how their results enter an authority-envelope view of a trajectory, are bounded against thresholds fixed in advance, and connect to controls and consequences. Production machine-learning practice has long recognized monitoring and readiness rubrics \citep{sculley2015,breck2017}, internal auditing defines end-to-end accountability \citep{raji2020}, and incident databases support learning from failures \citep{mcgregor2021}. \RBED{} adds consequence-linked monitoring tiers, an independent test that monitors work, and a gated loop from incident to new scenarios and to invalidation of evidence.

\paragraph{Consequence-aware risk and loss modeling.} Risk as scenarios with likelihoods and consequences is classical \citep{kaplan1981}; coherent and tail measures are standard \citep{artzner1999,rockafellar2000}; dependence bounds are due to Fr\'echet \citep{frechet1951}. Srivastava and Sah decompose expected loss from automated systems into failure, propagation to harm given automation, and severity, and develop elasticity and optimal-oversight results \citep{srivastava2026}; Section~\ref{sec:risk} relates it to the present work.

\begin{table}[!tbp]
\centering\footnotesize
\caption{Positioning of \RBED{} relative to related approaches. ``Adds'' states what \RBED{} proposes beyond the approach and is not a claim of empirical superiority.}
\label{tab:related}
\begin{tabular}{@{}L{3.0cm}L{4.4cm}L{7.6cm}@{}}
\toprule
Approach & Typical object and output & Relationship to \RBED{} \\
\midrule
AI RMF and Generative AI Profile \citep{nist_rmf,nist_genai} & Voluntary risk-management outcomes and actions & \RBED{} is a method for some of the evidence the outcomes call for; it creates no organizational governance and implies no conformity. \\
ISO/IEC 42001 \citep{iso42001} & Certifiable management system & Complementary; \RBED{} supplies system-level evidence, not a management system. \\
Model and sociotechnical evaluation \citep{liang2023helm,shevlane2023,weidinger2023} & Model or layered evaluation results & \RBED{} adds a bounded, dated evidence record and a rule-governed link to authorization. \\
Safety and assurance cases \citep{clymer2024} & Structured safety argument & \RBED{} fixes record fields, statuses and rules so that the argument is checkable for discipline. \\
Model risk management \citep{sr117,sr262} & Validation, monitoring and change governance for models & \RBED{} extends independent challenge and revalidation to configured, including agentic, systems with controls. \\
Agent benchmarks and red teaming \citep{yao2024tau,debenedetti2024,ruan2024} & Scenario suites and attack results & These are test sources; \RBED{} adds authority-envelope metrics, control-effectiveness evidence and consequence linkage. \\
Monitoring practice and incident databases \citep{sculley2015,breck2017,mcgregor2021} & Signals, rubrics, incident records & \RBED{} adds consequence-linked tiers, independent monitor validation and a gated learning loop. \\
Failure-mediated loss decomposition \citep{srivastava2026} & Theory of automation risk and oversight & \RBED{} reuses the decomposition and adds pathways, controls, evidence and authorization rules. \\
\bottomrule
\end{tabular}
\end{table}

\section{Problem Formulation and Design Requirements}\label{sec:problem}
\paragraph{Setting.} A \emph{configured system} $A$ is a model together with the prompts, tools, permissions, retrieval sources, memory, workflow, data, deployment controls and operating context in which it runs. Evidence is gathered about $A$ at a point in time. A \emph{system claim} is an observable acceptance boundary on $A$, quantitative or categorical, for example ``a server-side check rejects over-limit refunds regardless of model content.'' A \emph{consequence} is a harm of a stated kind to a stated party, with severity, reach, duration and reversibility. A \emph{decision} is an authorization: the disposition for $A$, by a named authority, for a bounded period, with conditions. The problem is to make the path from claims to decisions explicit, checkable and time-bounded.

\paragraph{Why the three gaps are distinct.} The evaluation gap concerns the object evaluated: configuration changes what a model can do and what happens after it errs. The consequence gap concerns the event evaluated: a failure $F$ is not a harm $H$, since a failure can be stopped before it reaches anyone, and harm can occur with no failure, through intended behavior or misuse. The evidence-to-decision gap concerns inference: complete test results do not say whether to deploy, how long a conclusion may be relied on, or what to do when a component changes.

\paragraph{Design requirements.} We derive eight requirements from the gaps (evaluation E, consequence C, evidence-to-decision D). They are design requirements, not theorems; whether meeting them improves decisions is the empirical question of Section~\ref{sec:discussion}.
\begin{enumerate}[label=\textbf{DR\arabic*.},leftmargin=2.6em]
  \item (E) Evaluate the configured system, including tools, permissions and controls, not the model alone: architecture and record (Section~\ref{sec:arch}).
  \item (C) Keep failure, harm and consequence distinct and cover harm without failure: pathways and identity (Section~\ref{sec:risk}).
  \item (C) Direct evaluation and monitoring effort by consequence, not frequency alone: register and tiers (Sections~\ref{sec:arch}, \ref{sec:cse}).
  \item (D) State what evidence supports a claim, for what population, versions and window, with uncertainty and independence: statuses and I0 to I3 (Section~\ref{sec:evidence}).
  \item (E, D) Credit a safeguard only when shown to work under realistic failure: assurance ladder and rule 6 (Section~\ref{sec:evidence}).
  \item (D) Make the decision traceable and rule-governed, never treating unresolved as safe: dispositions and eight rules (Section~\ref{sec:evidence}).
  \item (D) Make evidence expire and respond in proportion to change: Evidence Validity Register (Section~\ref{sec:cse}).
  \item (C, D) Learn from failures and test the monitors that detect them: monitoring validation and learning loop (Section~\ref{sec:cse}).
\end{enumerate}

\paragraph{Non-goals.} \RBED{} produces no composite risk score, does not price non-monetary harms for summation with financial ones, does not perform security testing, privacy impact assessment or legal analysis, and does not create organizational authority. The thirteen dimensions are diagnostic categories and are never scored, weighted or multiplied.

\section{\RBED{} Architecture}\label{sec:arch}
\begin{figure}[!t]
\begingroup\hyphenpenalty=10000 \exhyphenpenalty=10000 \centering\resizebox{0.84\linewidth}{!}{% Figure 2: R-BED triangle architecture. Geometry follows the book figure (same coordinates, 6+4+3 spacing).
% All positions derive from named coordinates; the triangle, circle, apex label and vertex labels share the axis x = 0.
\begin{tikzpicture}[font=\sffamily\scriptsize]
\coordinate (BL)  at (-5.76, 7.30);
\coordinate (BR)  at ( 5.76, 7.30);
\coordinate (AP)  at ( 0.00,14.30);
\coordinate (CC)  at ( 0.00,10.50);   % circle centre, on the symmetry axis
\coordinate (EQ)  at ( 0.00, 6.72);
\coordinate (HDR) at ( 0.00,-0.20);
% header bar and group caption
\node[fill=darkBlue, text=white, font=\sffamily\bfseries\small, minimum width=11.52cm, minimum height=0.90cm] at (HDR) {SYSTEM-LEVEL AI EVALUATION $\cdot$ R-BED};
\node[font=\sffamily\bfseries\small, text=dgC] at (0,-0.95) {6 + 4 + 3 = 13 dimensions};
% thirteen columns (identical width and height)
\foreach \px/\pc/\pl in {
  -4.68/v1C/{A1 Data Foundations}, -3.96/v1C/{A2 System Design}, -3.24/v1C/{A3 Predictive Validity},
  -2.52/v1C/{A4 Fairness}, -1.80/v1C/{A5 Failure Analysis}, -1.08/v1C/{A6 Explainability},
   0.00/v2C/{B1 AI Authority}, 0.72/v2C/{B2 Human Mediation}, 1.44/v2C/{B3 Exposure Control}, 2.16/v2C/{B4 Harm Detection},
   3.24/v3C/{C1 Incremental Impact}, 3.96/v3C/{C2 Scale}, 4.68/v3C/{C3 Reg. and Reputational}}{
  \node[draw=\pc, fill=white, line width=1.0pt, minimum width=0.56cm, minimum height=4.60cm] at (\px,2.927) {};
  \node[fill=\pc, minimum width=0.56cm, minimum height=0.22cm] at (\px,0.397) {};
  \node[fill=\pc, minimum width=0.56cm, minimum height=0.22cm] at (\px,5.507) {};
  \node[rotate=90, anchor=west, inner sep=2pt, text=dgC, font=\sffamily\bfseries\fontsize{9pt}{10pt}\selectfont] at (\px,0.707) {\pl};}
\draw[dashed, dgC!55, line width=0.7pt] (-0.54,0.441) -- (-0.54,5.941);
\draw[dashed, dgC!55, line width=0.7pt] ( 2.70,0.441) -- ( 2.70,5.941);
% applicability strip and equation (same width as the column block)
\node[fill=tealC, text=white, font=\sffamily\bfseries\small, minimum width=11.52cm, minimum height=0.50cm] at (0,5.981)
  {TRADITIONAL ML \ \textbullet\ \ GENAI \ \textbullet\ \ RAG \ \textbullet\ \ AGENTIC AI};
\node[draw=dgC, line width=0.8pt, fill=dgC!6, rounded corners=2pt, font=\sffamily\bfseries\small, text=dgC, align=center, minimum width=9.0cm, minimum height=0.70cm, inner xsep=10pt, inner ysep=4pt] at (EQ)
  {$\mathbf{\mathbb{E}[L_F \mid A] \;=\; P(F \mid A)\;\times\;P(H \mid F,A)\;\times\;\mathbb{E}[S \mid H,F,A]}$};
% triangle
\draw[line width=1.8pt, dgC, line join=round] (BL)--(AP)--(BR)--cycle;
% vertex descriptions: left and right mirror each other about x = 0 and share a baseline height
\node[font=\sffamily\bfseries\normalsize, text=v1C, align=center, anchor=east, inner sep=4pt, text width=3.1cm] at (-5.80,8.00)
  {Technical\\ Failure\\ (6 dimensions)\\[3pt]\textcolor{dgC}{$\mathbf{P(F \mid A)}$}};
\node[font=\sffamily\bfseries\normalsize, text=v3C, align=center, anchor=west, inner sep=4pt, text width=3.1cm] at ( 5.80,8.00)
  {Consequence\\ and Impact\\ (3 dimensions)\\[3pt]\textcolor{dgC}{$\mathbf{\mathbb{E}[S \mid H,F,A]}$}};
\node[font=\sffamily\bfseries\normalsize, text=v2C, align=center, anchor=south, inner sep=4pt] at (0,14.35)
  {Deployment Controls (4 dimensions)\\[3pt]\textcolor{dgC}{$\mathbf{P(H \mid F,A)}$}};
% central circle, centred on the axis; label centred with a fixed text width
\node[circle, draw=dgC, fill=white, line width=2.2pt, minimum size=3.80cm, font=\sffamily\bfseries\normalsize, text=dgC, align=center, text width=2.85cm, inner sep=0pt] at (CC) {Connected\\ evidence and\\ controls};
% method line below the circle
\node[font=\sffamily\bfseries\scriptsize, text=evalindigo!85!black, align=center] at (0,8.30) {BUILD $\rightarrow$ BREAK $\rightarrow$ EVAL $\rightarrow$ IMPROVE};
\node[font=\sffamily\itshape\scriptsize, text=dgC!80, align=center] at (0,7.85) {See Section~\ref{sec:lifecycle}};
\end{tikzpicture}}\par\endgroup
\caption{The \RBED{} architecture: three non-substitutable vertices and thirteen diagnostic dimensions (V1 technical failure, A1 to A6; V2 deployment controls, B1 to B4; V3 consequence and impact, C1 to C3, where C3 is regulatory and reputational impact with components C3a and C3b). The triangle conceptually connects the vertices to the conditional failure-mediated expected-loss decomposition $\Ex[L_F\cnd A]$ shown, but a vertex is not identical to its factor in every respect: a control in V2 can also bound exposure and severity. Dimensions are diagnostic locations for evidence, not mathematical weights or scores. Harm can also arise from intended behavior or misuse without a technical failure (Figure~\ref{fig:funnel}). The five objectives are lenses, not dimensions.}
\label{fig:arch}
\end{figure}
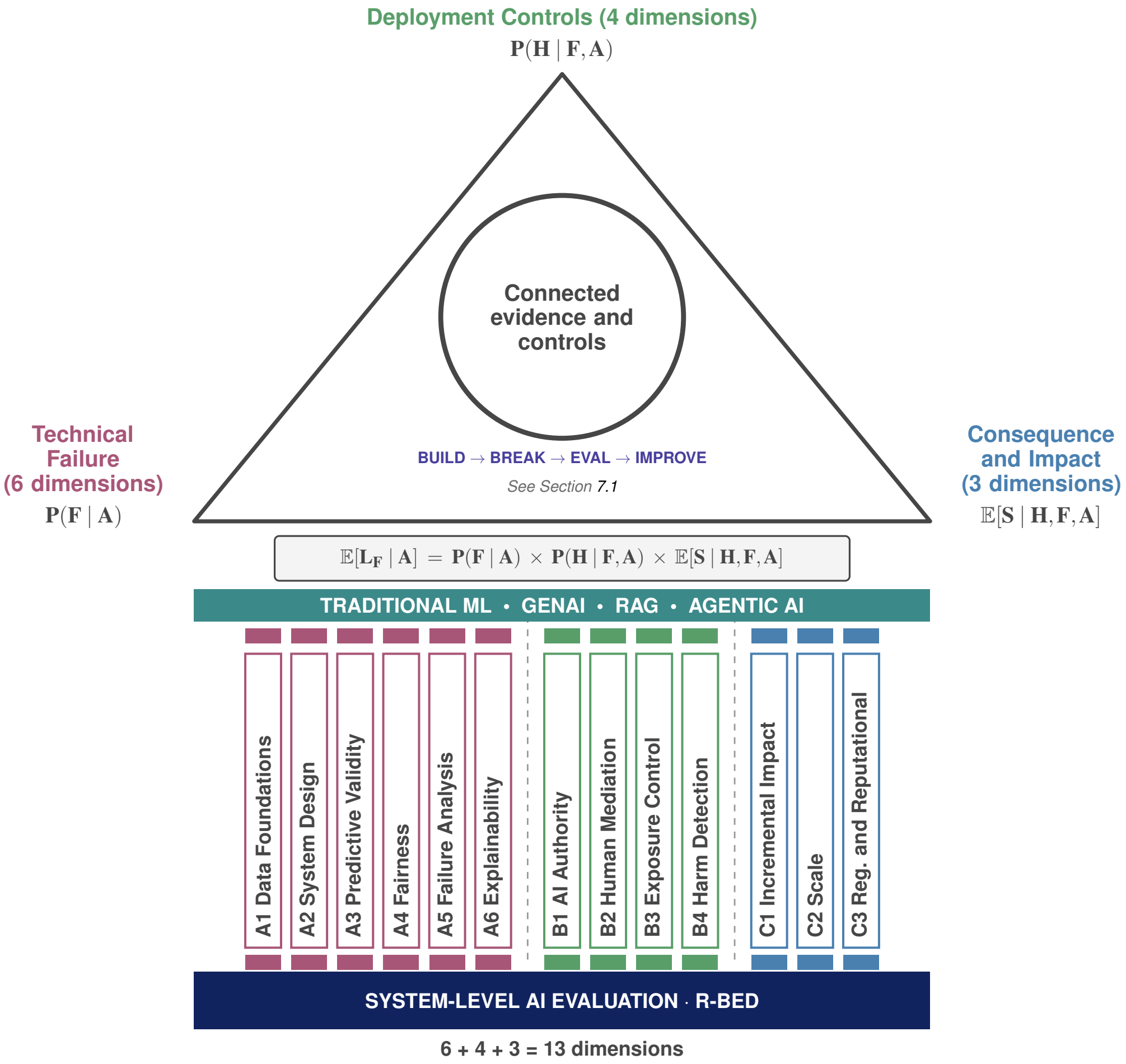

\textbf{Maturity: [E]} for the architecture, the objectives and the Consequence Register; the C3a/C3b components and the reputational category are proposed practice [P]. A \emph{dimension} is a diagnostic location for a claim and its evidence. It is not a standalone score, a level or a weight. The architecture is 6+4+3: three vertices with thirteen dimensions (Figure~\ref{fig:arch}).

\paragraph{Vertex 1: Technical Failure (A1 to A6).} A1 Data Foundations (source data, indexes and knowledge: authorized, suitable, current, representative); A2 System Design (architecture, interfaces, dependencies, failure containment); A3 Predictive Validity (does the configured system satisfy an explicit acceptance predicate); A4 Fairness (distribution of errors, benefits and harms across groups); A5 Failure Analysis (where, when and for whom claims break); A6 Explainability (can a reviewer reconstruct consequential behavior from traces, versions and lineage).

\paragraph{Vertex 2: Deployment Controls (B1 to B4).} A deployment control is a mechanism outside the model's own judgment that limits what the system can do, how far an effect spreads, or how quickly harm is noticed and stopped. B1 AI Authority: the inventory of actions the system can cause, each with a privilege level, scope and limit, every limit enforced by a mechanism that does not depend on model output. B2 Human Mediation: a human can effectively intervene, with the information, time, competence, authority and independence to detect and override. B3 Exposure Control: the magnitude, rate, scope and duration of any single failure or misuse are bounded before containment, and recoverable. B4 Harm Detection: harmful effects, not only technical errors, are detected with enough coverage, sensitivity and speed that a named owner can act before the consequence exceeds tolerance.

\paragraph{Vertex 3: Consequence and Impact (C1 to C3).} C1 Incremental Impact: the additional harm, financial and nonfinancial, attributable to an AI-influenced failure per affected exposure, against a stated counterfactual baseline and not against total business value or nominal exposure. C2 Scale: how widely consequences can occur (affected decisions, concentration, common-cause failure, duration). C3 Regulatory and Reputational Impact: assessed as two components, C3a regulatory and legal and C3b reputational and trust, which are components of C3, not additional dimensions, and are assessed separately. C3 can create financial consequences; where an amount is already counted in a direct loss estimate it is recorded once, where it is priced, and C3 cites it. C1 is the consequence per affected exposure, C2 how many exposures are affected and for how long, and C3 the institutional consequence that may follow; ordinal assessments are never combined by multiplication.

\paragraph{Consequence Register.} Consequences are recorded per system in twelve categories (physical safety, security, privacy, financial, legal and regulatory, discrimination and fairness, psychological, dignity and autonomy, operational continuity, environmental, societal and institutional, reputational and trust). Each entry records the scenario and pathway, affected parties, severity, reach, duration, reversibility, the tolerance or prohibition, and the claims it bears on. A consequence is \emph{high-consequence} if its severity is high or severe, or it is irreversible. The classification follows the \emph{inherent} consequence and is not lowered because a control exists; it triggers stricter rules (Section~\ref{sec:evidence}). A categorical prohibition cannot be exchanged for business benefit, and non-monetary categories are not converted to prices to be summed with financial ones.

\paragraph{Cross-cutting objectives.} Safety, security, privacy, robustness and resilience are lenses: questions asked of the thirteen dimensions, each with primary locations of evidence (for example, resilience is primarily located in B2, B3, B4, C1 and C2). Treating them as lenses preserves 6+4+3 and avoids additional ``dimensions'' that would invite scoring. For an in-scope security or privacy objective on a high-consequence system, the record must point to the specialized assessment; \RBED{} does not perform or substitute for it.

\paragraph{Evidence perspectives and workflow.} Evidence is component evidence (tests of models, tool wrappers, permission checks and monitors), trajectory evidence (the ordered record of retrieval, prompts, tool arguments, approvals and state transitions) or outcome evidence (consequences for users, records, money and rights), each informing several vertices. An eight-step workflow (context, claims, pathways and register, evidence design, break and characterize, challenge controls, evaluate and decide, fix, verify and renew) organizes the work.

\section{Consequence-Aware Risk Formulation}\label{sec:risk}
\textbf{Maturity:} the conditional identity is [E]; the extensions, the stressed-loss requirement and the exposure bookkeeping are [P]. None is a validated forecasting model.

\subsection{The failure-mediated identity}\label{sec:identity}
Fix one explicitly defined failure-mediated pathway, one exposure unit (a request, session, transaction or decision) and a configured system $A$. Let $F$ be the event that a failure of a stated kind occurs in one exposure, $H$ the event that this failure results in harm of a stated category \emph{after} the deployment controls act, and $S\ge 0$ a quantitative severity of that harm on a ratio scale defined for the category (currency where priced, or another measured unit such as hours of outage or persons affected). The identity below, and any expectation or sum built from it, applies only to categories with such an $S$. Ordinal or non-monetary consequences (for example regulatory or reputational effects in the Consequence Register) are not coded as numbers and averaged; they are assessed separately, for instance as the probability $\Prob(F\cnd A)\Prob(H\cnd F,A)\Prob(\text{level}\ge k\cnd H,F,A)$ of reaching a stated level $k$, or by category-specific judgment with evidence. Define the failure-attributable loss $L_F = S\,\mathbf 1\{F\}\,\mathbf 1\{H\}$. Then, by the chain rule and conditional expectation,
\begin{equation}\label{eq:identity}
\Ex[L_F\cnd A] \;=\; \Prob(F\cnd A)\;\Prob(H\cnd F,A)\;\Ex[S\cnd H,F,A],
\end{equation}
where $\Ex[S\cnd H,F,A]$ is defined only when $\Prob(H\cap F\cnd A)>0$. Units: $\Prob$ factors are dimensionless probabilities per exposure unit, and $\Ex[L_F\cnd A]$ is in the units of $S$ per exposure unit over a stated horizon. \emph{Assumptions:} the pathway, exposure unit, horizon and consequence definition are fixed; the probabilities and expectations exist; each factor is conditioned on the preceding events as written. \emph{Limits:} Equation~\eqref{eq:identity} is a conditional identity under these definitions. It is not a validated loss forecasting model, a universal risk score, or a claim that non-monetary harms can be priced. Each factor needs its own evidence: $\Prob(F\cnd A)$ requires failure evidence under a stated operating distribution; $\Prob(H\cnd F,A)$ requires challenge tests conditional on failure, which is where deployment controls enter; and $\Ex[S\cnd H,F,A]$ requires consequence evidence. If a conditioning event has zero probability, the factor needs separate treatment rather than a numerical reading. Dependencies, heavy tails, underreporting, shared-cause failures, unobserved pathways and overlapping events must be disclosed.

\paragraph{Relation to prior work.} The decomposition into failure, propagation to harm and severity is not new here: it is the central identity of Srivastava and Sah \citep{srivastava2026}, who write $\Ex[\text{Loss}] = \Prob(F)\,\Prob(H\cnd F,A)\,\Ex[S\cnd H]$ with $A$ an automation level, assume that harm requires failure ($\Prob(H\cnd\neg F,A)=0$) and that failure probability does not depend on automation, and develop elasticity, efficient-frontier and resource-allocation results, which are theoretical with empirical validation left to future work. Equation~\eqref{eq:identity} differs in three respects. It conditions all factors on the configured system $A$ (so $\Prob(F\cnd A)$ is not assumed independent of the configuration), it conditions severity on the failure and harm events as well, and it is embedded in a framework that treats harm without failure and that links each factor to a kind of evidence. We do not extend the prior mathematical results.

\subsection{Pathways, inherent and residual risk}\label{sec:pathways}
\begin{figure}[!tbp]
\begingroup\hyphenpenalty=10000 \exhyphenpenalty=10000 \centering\resizebox{0.74\linewidth}{!}{\begin{tikzpicture}[font=\sffamily\footnotesize]
% funnel silhouette: three non-substitutable stages; widths are illustrative only
\draw[draw=v1C, fill=v1C!12, line width=0.9pt] (-4.4,3.6)--(4.4,3.6)--(3.6,2.4)--(-3.6,2.4)--cycle;
\draw[draw=v2C, fill=v2C!12, line width=0.9pt] (-3.6,2.4)--(3.6,2.4)--(2.8,1.2)--(-2.8,1.2)--cycle;
\draw[draw=v3C, fill=v3C!12, line width=0.9pt] (-2.8,1.2)--(2.8,1.2)--(2.0,0)--(-2.0,0)--cycle;
\node[align=center, font=\sffamily\bfseries\footnotesize] at (0,3.0) {Technical Failure (A1 to A6)\\[-1pt]\mdseries\scriptsize widest: most places a claim can break};
\node[align=center, font=\sffamily\bfseries\footnotesize] at (0,1.8) {Deployment Controls (B1 to B4)\\[-1pt]\mdseries\scriptsize limit how much reaches the world};
\node[align=center, font=\sffamily\bfseries\footnotesize] at (0,0.6) {Consequence and Impact\\[-1pt]\mdseries\scriptsize (C1 to C3)};
\node[box, text width=24mm, anchor=east] (nn) at (-4.1,1.8) {Intended behavior $N$\\ \scriptsize harm without failure};
\node[box, text width=24mm, anchor=west] (mm) at (4.1,1.8) {Misuse $M$\\ \scriptsize attempt and success};
\draw[dflow] (nn.east)--(-3.2,1.8); \draw[dflow] (mm.west)--(3.2,1.8);
\end{tikzpicture}}\par\endgroup
\caption{The architecture as a risk-narrowing pathway (illustrative). Technical failure is drawn widest because it contains the most dimensions and the most places where a claim can break; deployment controls determine how much of a failure reaches the world, and consequence and impact asks how much it matters. The widths illustrate the narrowing logic only; they do not indicate probability, weight or importance, and the three vertices remain non-substitutable. Harm can also arise from intended behavior $N$ or misuse $M$ without a technical failure. Depending on the defined pathway and configuration $A$, a control may act on $\Prob(H\cnd F,A)$, on these pathways, or on $\Prob(F\cnd A)$ itself; the figure shows the typical case. Pathway losses are kept separate and overlap is bounded (Section~\ref{sec:bounds}).}
\label{fig:funnel}
\end{figure}
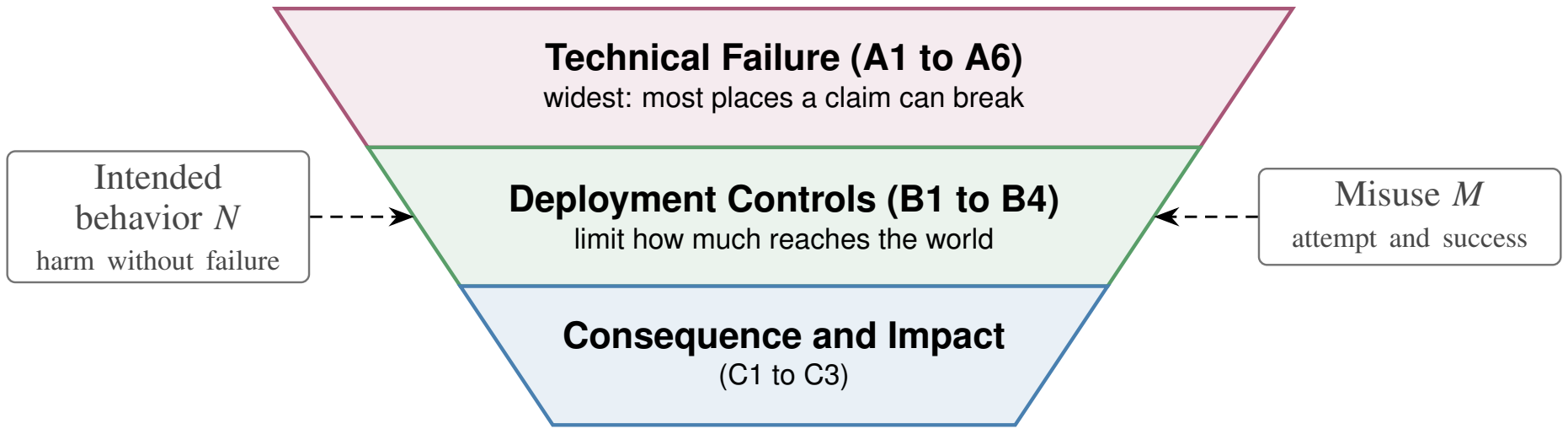

Three pathways are kept separate (Figure~\ref{fig:funnel}). \emph{Failure-mediated} harm is Equation~\eqref{eq:identity}. \emph{Intended-behavior} harm $N$ is caused by behavior that meets every claim, so no failure factor appears: $\Ex[L_N\cnd A]=\Prob(N\cnd A)\,\Ex[S\cnd N,A]$. A system that passes every claim can still cause harm, and this pathway requires consequence evidence for correct behavior. \emph{Misuse} $M$ is harm from a person or agent using the system for an unintended purpose: $\Ex[L_M\cnd A]=\Prob(\text{attempt})\,\Prob(\text{success}\cnd\text{attempt},\text{controls})\,\Ex[S\cnd\text{success}]$. Attempt probability is not stationary, since it responds to the system and its controls, so results hold for the tested effort and version.

Three terms are used in a fixed order. \emph{Materiality} is the significance of the purpose, exposure and potential consequences judged before testing and before controls; it prioritizes which claims are evaluated, at what depth and independence. \emph{Inherent risk} is the consequence potential of a pathway under the intended authority and exposure with deployment controls not yet credited; it fixes the high-consequence classification. \emph{Residual risk} is what remains for a specified configuration after the controls that have been tested and credited with evidence: the factors of Equation~\eqref{eq:identity} estimated with those controls in place, recorded as an upper bound or an explicit abstention, never silently zero. Residual risk, with its uncertainty, is what the approving authority compares with the stated tolerance. A strong control changes the residual risk and the evidence required to keep it, not the inherent classification.

\subsection{Exposure populations and incremental loss}\label{sec:populations}
Five quantities are kept apart, and every rate or loss names its exposure unit and denominator: \emph{total business exposure} (everything the activity puts at risk, AI or not); \emph{AI-influenced exposure} (where the system's output influenced a decision); \emph{affected} decisions (touched by a failure $F$, whether or not harm followed); \emph{harmed} decisions (where $H$ occurred); and \emph{incremental loss}, $L_{\text{incr}} = L_{\text{with}} - L_{\text{baseline}}$ against a stated counterfactual, which can be negative and needs a supporting design for a causal reading. Nominal exposure is not a loss estimate. For units or pathways $i=1,\dots,m$ with finite expectations and no loss counted twice, $\Ex[L_{\text{total}}]=\sum_i \Ex[L_i]$ by linearity, without independence; dependence matters for everything beyond the mean.

\subsection{Expected, stressed and tail loss}\label{sec:tail}
Expected loss $\Ex[L]$ is an average and can be small while a severe outcome is possible. \emph{Stressed loss} is the loss under a stated severe-but-plausible scenario, with its assumptions listed and no probability required: the question is whether the system and the organization can withstand it. A \emph{tail measure}, such as $\mathrm{VaR}_\alpha$ or $\mathrm{ES}_\alpha=\Ex[L\cnd L\ge \mathrm{VaR}_\alpha]$ \citep{rockafellar2000,artzner1999}, is used only where a defensible loss distribution exists; otherwise stress scenarios and the bounds below are reported and no tail figure is invented. A stress scenario covers common-cause failure of a shared component, correlated exposures, concentration and potentially catastrophic consequences. For each high-consequence Register entry, and for any system granted authority to act at scale, the record holds at least one stress scenario; without an assessment the stressed-tolerance claim is Unresolved. Favorable expected-loss evidence does not stand in for it.

\subsection{Bounds for dependence, overlap and thin evidence}\label{sec:bounds}
Independence of controls or pathways is never assumed without evidence. Table~\ref{tab:bounds} lists the bounds used. Each is a statement about intervals, not an estimate.

\begin{table}[!tbp]
\centering\footnotesize
\caption{Bounds used when evidence is thin or dependence is unknown. Further rows (independent controls, Bonferroni combination, exceedance over exposure and abstention) and numerical checks are in Appendix~\ref{app:details}; all examples are verified by the script of Appendix~\ref{app:repro}.}
\label{tab:bounds}
\begin{tabular}{@{}L{3.1cm}L{6.2cm}L{5.4cm}@{}}
\toprule
Item & Statement & Assumption, limit and check \\
\midrule
Union of pathways & For nonnegative losses with union loss equal to the pointwise maximum, $\max_i \Ex[L_i]\le\Ex[L_\cup]\le\sum_i\Ex[L_i]$. & Bounds, not an estimate; the sum is the safe bound when overlap is unknown. \\
Dependent controls (Fr\'echet \citep{frechet1951}) & For failure probabilities $p_a,p_b$, $\max(0,p_a+p_b-1)\le\Prob(\text{both fail})\le\min(p_a,p_b)$. & Use unless independence is demonstrated. For $p_a=0.1$, $p_b=0.2$ the interval is $[0,0.1]$ (checked). \\
Zero-failure trial counts & Zero failures in $n$ trials give a one-sided upper bound $1-(1-c)^{1/n}$. At 95\% confidence, $29$, $299$ and $2{,}995$ clean trials bound the rate below $0.10$, $0.01$ and $0.001$. & Counts checked. Assumes independent trials from the operating distribution; a bound is not a measurement. \\
Control effectiveness by contrast & Compare $\Prob(H\cnd F,\text{control off})$ with $\Prob(H\cnd F,\text{control on})$ under otherwise identical tests; report the upper bound with the control over the lower bound without. & Causal only if the arms differ solely in the control. Hypothetical 12/40 without and 1/80 with give a conservative ratio of $0.41$ (checked). \\
\bottomrule
\end{tabular}
\end{table}

A hypothetical three-configuration comparison with invented values (Appendix~\ref{app:details}) shows that containment acts on $\Prob(H\cnd F,A)$ and not on $\Prob(F\cnd A)$, and that a small expected value never makes a prohibited or irreversible harm acceptable.

\section{Evidence Quality and Deployment-Control Assurance}\label{sec:evidence}
\textbf{Maturity:} evidence statuses, independence levels, dispositions and the eight rules are [E]; the control assurance ladder is [P] (it builds on [E] independence levels and control challenge).

\subsection{Three separate records}\label{sec:threerecords}
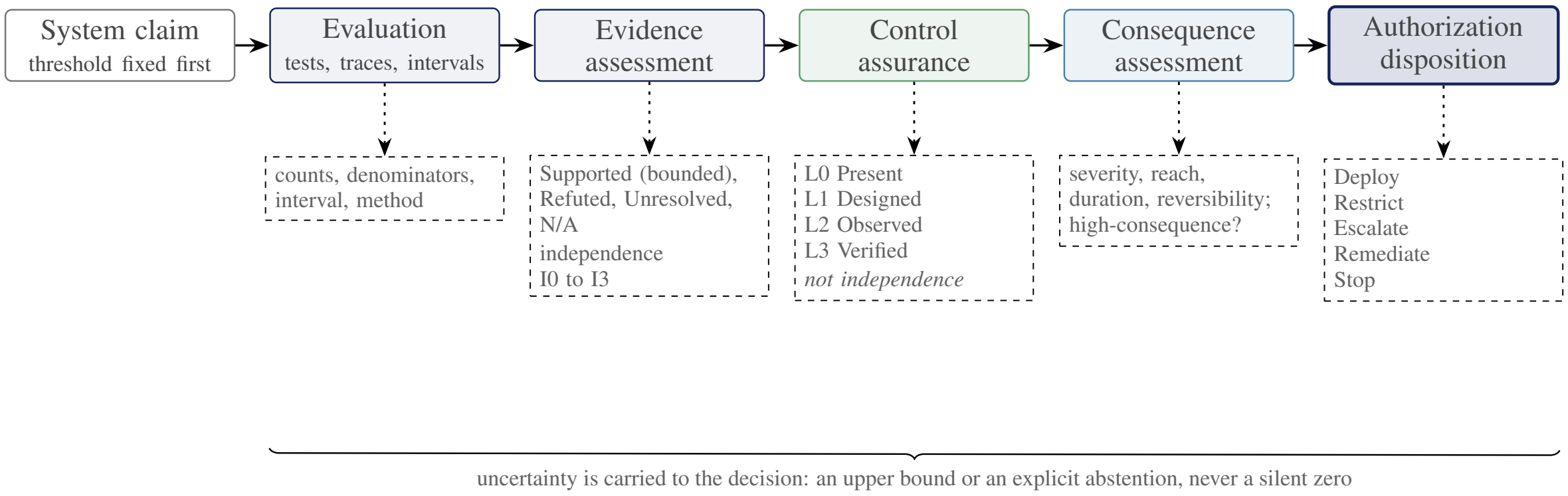
\begin{figure}[!htb]
\begingroup\hyphenpenalty=10000 \exhyphenpenalty=10000 \centering\resizebox{0.94\linewidth}{!}{\begin{tikzpicture}
  \def\w{23mm}
  \node[box, text width=\w] (a) at (0,0) {System claim\\ \scriptsize threshold fixed first};
  \node[proc, text width=\w] (b) at (29mm,0) {Evaluation\\ \scriptsize tests, traces, intervals};
  \node[proc, text width=\w] (c) at (58mm,0) {Evidence assessment};
  \node[ctl, text width=\w] (d) at (87mm,0) {Control assurance};
  \node[harm, text width=\w] (e) at (116mm,0) {Consequence assessment};
  \node[dec, text width=\w] (f) at (145mm,0) {Authorization disposition};
  \foreach \x/\y in {a/b,b/c,c/d,d/e,e/f} \draw[flow] (\x)--(\y);
  % details
  \node[note, below=8mm of c, text width=24mm, draw, dash pattern=on 2pt off 2pt, inner sep=3pt, align=left] (cs)
    {Supported (bounded), Refuted, Unresolved, N/A\\[1pt] independence\\ I0 to I3};
  \node[note, below=8mm of d, text width=24mm, draw, dash pattern=on 2pt off 2pt, inner sep=3pt, align=left] (ds)
    {L0 Present\\ L1 Designed\\ L2 Observed\\ L3 Verified\\[1pt] \itshape not independence};
  \node[note, below=8mm of f, text width=24mm, draw, dash pattern=on 2pt off 2pt, inner sep=3pt, align=left] (fs)
    {Deploy\\ Restrict\\ Escalate\\ Remediate\\ Stop};
  \node[note, below=8mm of e, text width=24mm, draw, dash pattern=on 2pt off 2pt, inner sep=3pt, align=left] (es)
    {severity, reach, duration, reversibility; high-consequence?};
  \node[note, below=8mm of b, text width=24mm, draw, dash pattern=on 2pt off 2pt, inner sep=3pt, align=left] (bs)
    {counts, denominators, interval, method};
  \foreach \x/\y in {b/bs,c/cs,d/ds,e/es,f/fs} \draw[dotflow] (\x.south)--(\y.north);
  % uncertainty bar
  \draw[decorate, decoration={brace, mirror, amplitude=3pt}, line width=0.5pt] ([yshift=-40mm]b.south west) -- node[below=3pt, note] {uncertainty is carried to the decision: an upper bound or an explicit abstention, never a silent zero} ([yshift=-40mm]f.south east);
\end{tikzpicture}}\par\endgroup
\caption{Evidence-to-authorization chain. Each stage is a separate record. Evidence is bounded by status and independence; control assurance (L0 to L3) is a different axis from independence; uncertainty is carried to the decision as an upper bound or an explicit abstention.}
\label{fig:chain}
\end{figure}

\RBED{} keeps three records distinct so that one cannot silently stand for another (Figure~\ref{fig:chain}). The \emph{test result} holds counts, denominators, intervals and traces. The \emph{evidence assessment} holds the status of each claim and why. The \emph{deployment decision} holds the disposition and authorization. A passing test does not by itself change a claim's status, and a status does not by itself authorize deployment.

\subsection{Bounded evidence statuses and independence}\label{sec:statuses}
Each material claim has exactly one of four statuses. \textbf{Supported (bounded)}: adequate evidence supports the claim only for the specified population, versions, time window and conditions. For a claim measured as a rate, the acceptance threshold (the maximum tolerable rate) is fixed \emph{before} the test, and the claim is Supported only if the upper confidence bound of the result is at or below it; a result whose upper bound exceeds the threshold leaves the claim Unresolved. \textbf{Refuted}: admissible evidence shows violation of a required condition or control. \textbf{Unresolved}: crucial evidence, coverage, validity, independence or uncertainty is inadequate; unresolved is not synonymous with safe. \textbf{Not applicable}: a documented, reviewable justification explains why a dimension or claim does not apply. A bound is not a measurement: with zero failures at 95\% confidence, 29 trials suffice against a threshold of 0.10 but 2{,}995 are needed against 0.001 (Table~\ref{tab:bounds}), so the threshold is chosen from risk tolerance first.

Independence describes how far the producer of evidence is separated from those who build or operate the system. \textbf{I0}: self-assessed by the building or operating team. \textbf{I1}: a separate team in the same reporting line. \textbf{I2}: an independent function within the organization with its own reporting line. \textbf{I3}: an external party with no commercial dependence on the outcome. Independence qualifies evidence; it does not make it correct, and higher independence is a design choice proportional to consequence.

\subsection{Control assurance ladder}\label{sec:ladder}
A safeguard on the page is not a safeguard in operation. Four levels, which are cumulative, separate what is often reported as one: \textbf{L0 Present} (the control exists and is recorded); \textbf{L1 Designed} (its design is reviewed: enforced outside the model's judgment, no model-controlled path around it, fails safe); \textbf{L2 Observed} (measured effectiveness under stated challenge conditions meets thresholds fixed in advance, each measure on its conservative bound, upper for failure rates and lower for success rates); \textbf{L3 Verified} (an independent party, I2 or I3, verified the effectiveness evidence). The ladder describes the \emph{assurance of the control}, whereas independence describes the \emph{producer of evidence}; they are different axes. A control that is present, reviewed and measured but whose measure misses its threshold stays at L1, and a control a team measured on itself (I0) is L2 at most. A control credited for a claim tied to a high-consequence entry must reach L2, and L3 is expected. Eight control types (tool authorization, least privilege and policy enforcement in B1; human approval and escalation in B2; containment and recovery in B3; monitoring and alerting and logging in B4) each fall in an existing dimension.

\subsection{Dispositions and authorization rules}\label{sec:rules}
The canonical dispositions are five. \textbf{Deploy}: operate within the stated boundary on the stated evidence, with a current authorization record. \textbf{Restrict}: operate only within explicit, bounded conditions (scope, exposure, duration, population) while evidence is incomplete, with a current authorization record that states the conditions. \textbf{Escalate}: the decision exceeds the evaluator's or approver's authority, or the evidence state requires a higher authority; no operation beyond what is already authorized. \textbf{Remediate}: do not extend operation; fix the identified deficiency and retest independently. \textbf{Stop}: withdraw the system or capability, by suspension pending resolution or by retirement. ``Continue'' is the renewal of a Deploy authorization with a new record; it is not a sixth disposition. Eight rules are the conditions that a Deploy or Restrict must meet.

\begin{table}[!tbp]
\centering\footnotesize
\caption{The eight authorization rules (summarized from the Public Framework). They are checks on the discipline of a record; satisfying them does not show that the system is safe.}
\label{tab:rules}
\begin{tabular}{@{}L{0.9cm}L{3.7cm}L{10.2cm}@{}}
\toprule
Rule & Name & Condition \\
\midrule
1 & Named, current authorization & A Deploy or Restrict has an authorization record with an authorizer, a date and an expiry. An expired authorization is no authorization. \\
2 & Invalidation and monitoring & The record states the conditions that invalidate it and the monitoring that detects them. \\
3 & No refuted claim; prohibitions supported; thresholds fixed first & Deploy is not permitted with a refuted material claim, or with a categorical prohibition that is refuted or unresolved. A rate claim is Supported only against a threshold fixed before testing and only if the result's upper bound is at or below it. \\
4 & Unresolved is not safe & Deploy is not permitted while a claim tied to a high-consequence Register entry is unresolved. \\
5 & Restrict is bounded and must prevent the pathway & A Restrict states scope, exposure limit, duration and population. Where a refuted or unresolved claim is tied to a high-consequence entry, it must also name a credited control, enforced outside the model and with effectiveness evidence, that prevents that pathway from reaching affected people. An exposure percentage limits scale; it is not a restriction on the pathway. \\
6 & Controls credited only with evidence & A control on which the risk argument relies has effectiveness evidence; limits on authority are enforced outside the model. \\
7 & Independence and separation for high consequence & Where a high-consequence entry exists, evidence is not I0-only, the approving authority differs from the evaluation and business owners, and in-scope security and privacy objectives cite their specialized assessments. \\
8 & Residual risk is never silently zero & Residual risk is an upper bound or an explicit abstention. Transfer or acceptance of risk is not recorded as harm reduction. Exceptions expire. \\
\bottomrule
\end{tabular}
\end{table}

Any exception to a rule is recorded with a rationale, an approver and an expiry, and does not alter the evidence status of any claim. The rules check a record, not the world: a record that satisfies every rule can describe an unsafe system if the tests were poor, so the evaluation method itself (construct validity, sampling, coverage of rare and severe regions) must be reviewed before results are treated as decision-grade.

\section{Continuous System Evaluation}\label{sec:cse}
Continuous System Evaluation (CSE) answers a question that Section~\ref{sec:evidence} leaves open: for how long does a conclusion hold, and what happens when it stops holding? \textbf{Maturity:} the CSE protocol, the Evidence Validity Register, control challenge, agentic and oversight evaluation, monitoring tiers and the learning loop are proposed [P]; the benefit of the monitoring allocation is hypothesis [H] (H7). All cadences, thresholds and tables below are illustrative defaults to be set and justified by the adopting organization, not empirically derived values.

\subsection{Lifecycle and what continuous assurance means}\label{sec:lifecycle}
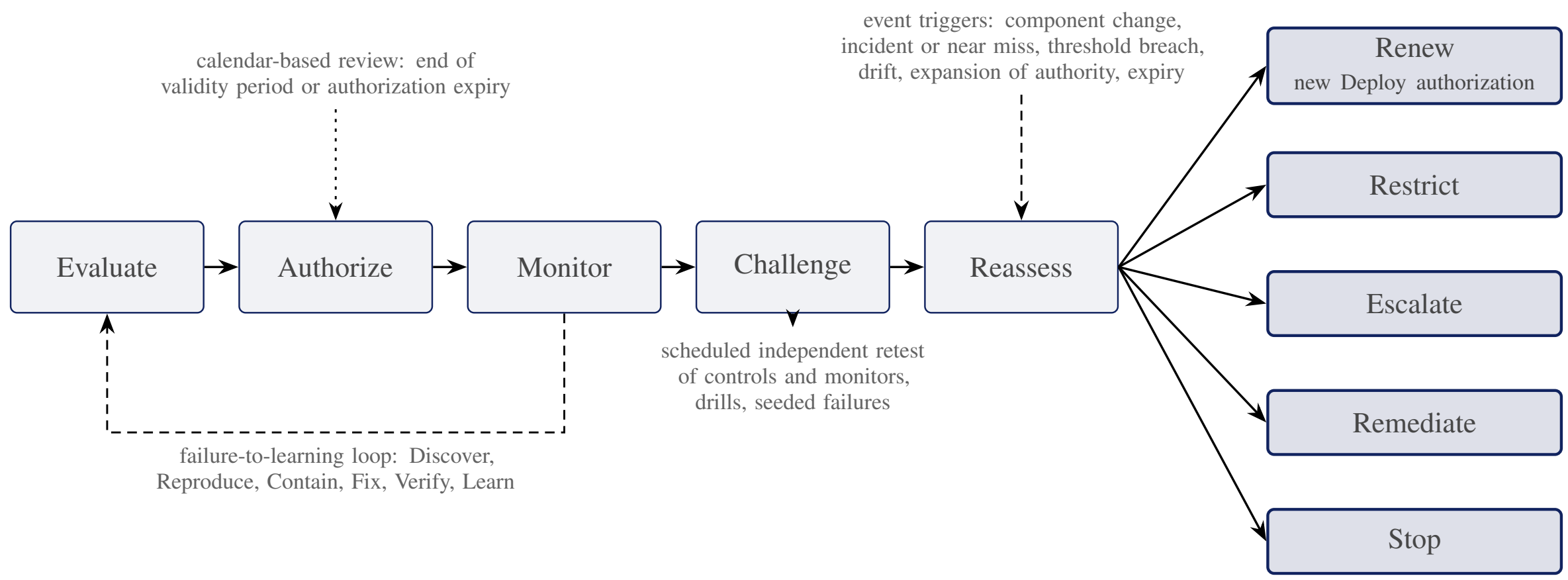
\begin{figure}[!tbp]
\begingroup\hyphenpenalty=10000 \exhyphenpenalty=10000 \centering\resizebox{0.94\linewidth}{!}{\begin{tikzpicture}
  \def\bw{19mm}
  \foreach \n/\x in {Evaluate/0, Authorize/25, Monitor/50, Challenge/75, Reassess/100}
    \node[proc, text width=\bw, minimum height=10mm] (\n) at (\x mm, 0) {\n};
  \foreach \a/\b in {Evaluate/Authorize, Authorize/Monitor, Monitor/Challenge, Challenge/Reassess} \draw[flow] (\a)--(\b);
  \node[dec, text width=30mm] (o1) at (143mm, 22mm) {Renew\\ \scriptsize new Deploy authorization};
  \node[dec, text width=30mm] (o2) at (143mm, 9mm) {Restrict};
  \node[dec, text width=30mm] (o3) at (143mm, -4mm) {Escalate};
  \node[dec, text width=30mm] (o4) at (143mm, -17mm) {Remediate};
  \node[dec, text width=30mm] (o5) at (143mm, -30mm) {Stop};
  \foreach \o in {o1,o2,o3,o4,o5} \draw[flow] (Reassess.east) -- (\o.west);
  \node[note, text width=50mm] (tr) at (100mm, 24mm) {event triggers: component change, incident or near miss, threshold breach, drift, expansion of authority, expiry};
  \draw[dflow] (tr.south) -- (Reassess.north);
  \node[note, text width=44mm] (cal) at (25mm, 21mm) {calendar-based review: end of validity period or authorization expiry};
  \draw[dotflow] (cal.south) -- (Authorize.north);
  \draw[dflow] (Monitor.south) -- ++(0,-13mm) -| (Evaluate.south);
  \node[note, text width=60mm, anchor=north] at (25mm,-18.5mm) {failure-to-learning loop: Discover, Reproduce, Contain, Fix, Verify, Learn};
  \node[note, text width=36mm, anchor=north] at (75mm,-7mm) {scheduled independent retest of controls and monitors, drills, seeded failures};
  \draw[dotflow] (Challenge.south) -- ++(0,-1.5mm);
\end{tikzpicture}}\par\endgroup
\caption{The CSE lifecycle. Event triggers and calendar review feed reassessment. The dispositions are the five canonical ones; renewal is a new Deploy authorization, not a sixth disposition. A failure-to-learning loop returns from monitoring to evaluation.}
\label{fig:cse}
\end{figure}

The lifecycle is Evaluate, Authorize, Monitor, Challenge and Reassess, followed by one of the canonical dispositions, with renewal being a new Deploy authorization (Figure~\ref{fig:cse}). Evaluation defines claims, thresholds and tests and records evidence with status, independence and validity; authorization issues a disposition with conditions, expiry, invalidation conditions and monitoring; monitoring collects consequence-linked signals at a tier cadence; challenge independently retests controls and monitors; reassessment checks evidence validity and authorization dependencies and then renews, restricts, escalates, remediates or stops.

\begin{figure}[!tbp]
\begingroup\hyphenpenalty=10000 \exhyphenpenalty=10000 \centering\resizebox{0.95\linewidth}{!}{% Build, Break, Eval, Improve: iterative claim verification (clockwise loop around the System evaluation core).
\begin{tikzpicture}[font=\sffamily\footnotesize]
\tikzset{stage/.style={box, font=\sffamily\footnotesize, draw=evalindigo, line width=0.9pt, text width=47mm, minimum height=15mm, inner sep=4pt, align=center, text=dgC},
         num/.style={circle, fill=evalindigo, text=white, font=\sffamily\bfseries\scriptsize, inner sep=1.2pt, minimum size=4.6mm}}
% core
\node[circle, draw=evalindigo, fill=evalindigo!10, line width=1.2pt, minimum size=21mm, align=center, font=\sffamily\bfseries\small, text=dgC, inner sep=1pt] (core) at (0,0) {System\\ evaluation};
\draw[dotted, evalindigo!60] (0,0) circle (1.62);
% stages
\node[stage, fill=evalindigo!6]  (b) at (0, 2.95) {{\bfseries\color{evalindigo!85!black}BUILD}\\[1pt]\scriptsize Define the system claim and deployment boundary};
\node[stage, fill=evalindigo!10] (k) at (5.6, 0) {{\bfseries\color{evalindigo!85!black}BREAK}\\[1pt]\scriptsize Seek disproof across failure regions, controls, authority and workflow};
\node[stage, fill=evalindigo!18] (e) at (0,-2.95) {{\bfseries\color{evalindigo!85!black}EVAL}\\[1pt]\scriptsize Synthesize the evidence, assign statuses, decide the disposition};
\node[stage, fill=evalindigo!10] (i) at (-5.6, 0) {{\bfseries\color{evalindigo!85!black}IMPROVE}\\[1pt]\scriptsize Change the system or its controls, then monitor; each change is a new claim that returns to BUILD};
\node[num] at (b.north east) {1}; \node[num] at (k.north east) {2}; \node[num] at (e.south east) {3}; \node[num] at (i.north east) {4};
% clockwise arcs between stages (radius 2.15)
\draw[cflow, evalindigo] (62:2.15) arc (62:28:2.15);
\draw[cflow, evalindigo] (-28:2.15) arc (-28:-62:2.15);
\draw[cflow, evalindigo] (-118:2.15) arc (-118:-152:2.15);
\draw[cflow, evalindigo] (152:2.15) arc (152:118:2.15);
% banner
\node[box, draw=evalindigo, fill=evalindigo!6, minimum width=11.4cm, minimum height=7mm, font=\sffamily\bfseries\small, text=evalindigo!85!black] at (0,-4.95) {Every intervention is a new claim: seek disproof, then verify the improvement.};
\end{tikzpicture}}\par\endgroup
\caption{The Build, Break, Eval, Improve loop (proposed, [P]). The four stages circle the system evaluation they all draw on, which holds connected evidence across the three vertices. A material improvement changes the evaluated system, so it creates a revised claim that must itself be built, broken and verified; new failure modes found while verifying return to BREAK.}
\label{fig:bbef}
\end{figure}
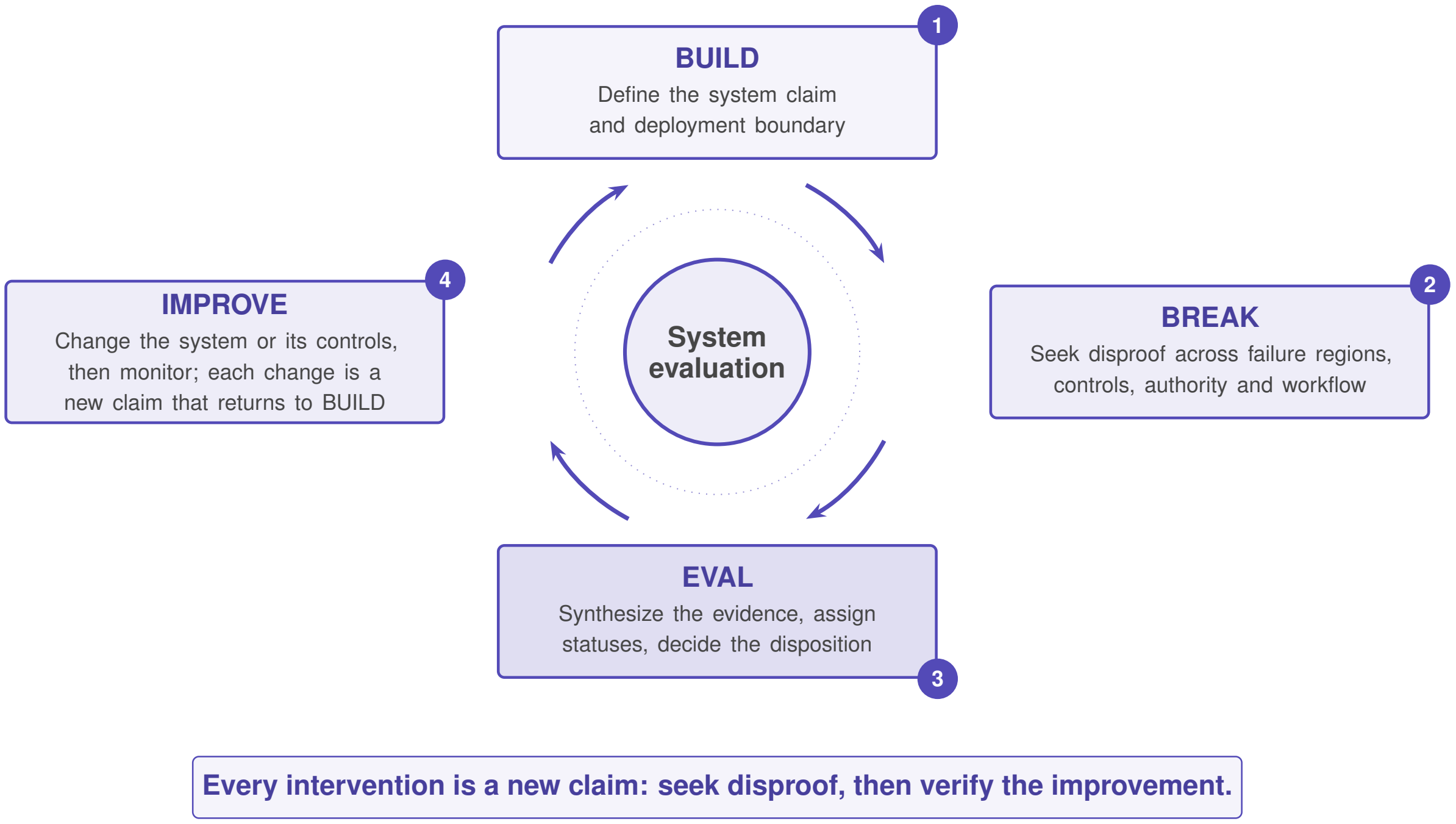

\paragraph{Build, Break, Eval, Improve.} Figure~\ref{fig:bbef} states the idea that \emph{every intervention is a new claim}. Restricting an agent's permissions, adding human review or refreshing data each asserts something new about how the system now behaves, and that assertion is not self-validating: it needs its own evidence, boundary and decision, so an improvement does not end the evaluation. The stages are not a strict waterfall. BREAK may force a claim to be restated (back to BUILD). IMPROVE covers the change and the monitoring afterwards, whose reassessment conditions trigger renewal. Because the change is itself a new claim, the loop returns to BUILD to state it, and verification of the repair is the loop running again (BREAK and EVAL), which may uncover a new failure region. EVAL is explicit because synthesis is where many evidence gaps are discovered. A safeguard is not verified by being built: the claim that a control works needs appropriate, independently reviewable effectiveness evidence, such as adversarial testing, formal or analytical verification, or other methods suited to the control, so the claim that a control holds meets the same evidentiary standard as the claim it protects.

\emph{Continuous assurance does not require continuously rerunning every evaluation test.} Four activities differ in cadence and in what they establish (Table~\ref{tab:activities}, Appendix~\ref{app:details}): calendar-based review, continuous monitoring, scheduled challenge and event-triggered reassessment. Evidence has a validity period, not a refresh rate: it is reviewed when its period ends or when a declared dependency changes, whichever comes first.

\subsection{Evidence Validity Register and proportional invalidation}\label{sec:evr}
Evidence supports a claim only for the configuration, conditions and period in which it was produced. The Evidence Validity Register (EVR) records, for each item that a claim or an authorization relies on: the claim, the supporting test and trace pointers, the \emph{system configuration pins} (model, prompts, tools, permissions, retrieval sources, memory, workflow, data, control and monitoring configuration), the tested operating conditions, the evidence timestamp, the owner, the uncertainty, a review-by date and optional hard expiry (the review-by date cannot precede the evidence date and the expiry cannot precede the review-by date), invalidation triggers (required for evidence on high-consequence claims), the required reassessment, and the authorization dependency with declared component dependencies.

A change does not invalidate everything. Each record declares the components on which it depends and how critical each dependency is. A change event names a component kind, its identity and a magnitude, and the effect on a record follows its declared dependency and the size of the change (Table~\ref{tab:impact}). Component kinds include model, prompt, tool, permission, retrieval source, memory, workflow, data, control configuration and monitoring configuration. The four impacts are \emph{no effect} (record the change), \emph{review} (a person confirms the evidence still holds), \emph{retest} (the evidence is not relied on for new decisions until the relevant test is rerun, and may support the existing authorization only until a stated retest deadline) and \emph{invalid} (the evidence no longer supports the claim).

\begin{table}[!tbp]
\centering\footnotesize
\caption{Default impact table (illustrative defaults; organizations may tighten them).}
\label{tab:impact}
\begin{tabular}{@{}lccc@{}}
\toprule
Declared dependency & Patch & Minor & Major \\
\midrule
Critical & Review & Retest & Invalid \\
Relevant & No effect & Review & Retest \\
Peripheral & No effect & No effect & Review \\
\bottomrule
\end{tabular}
\end{table}

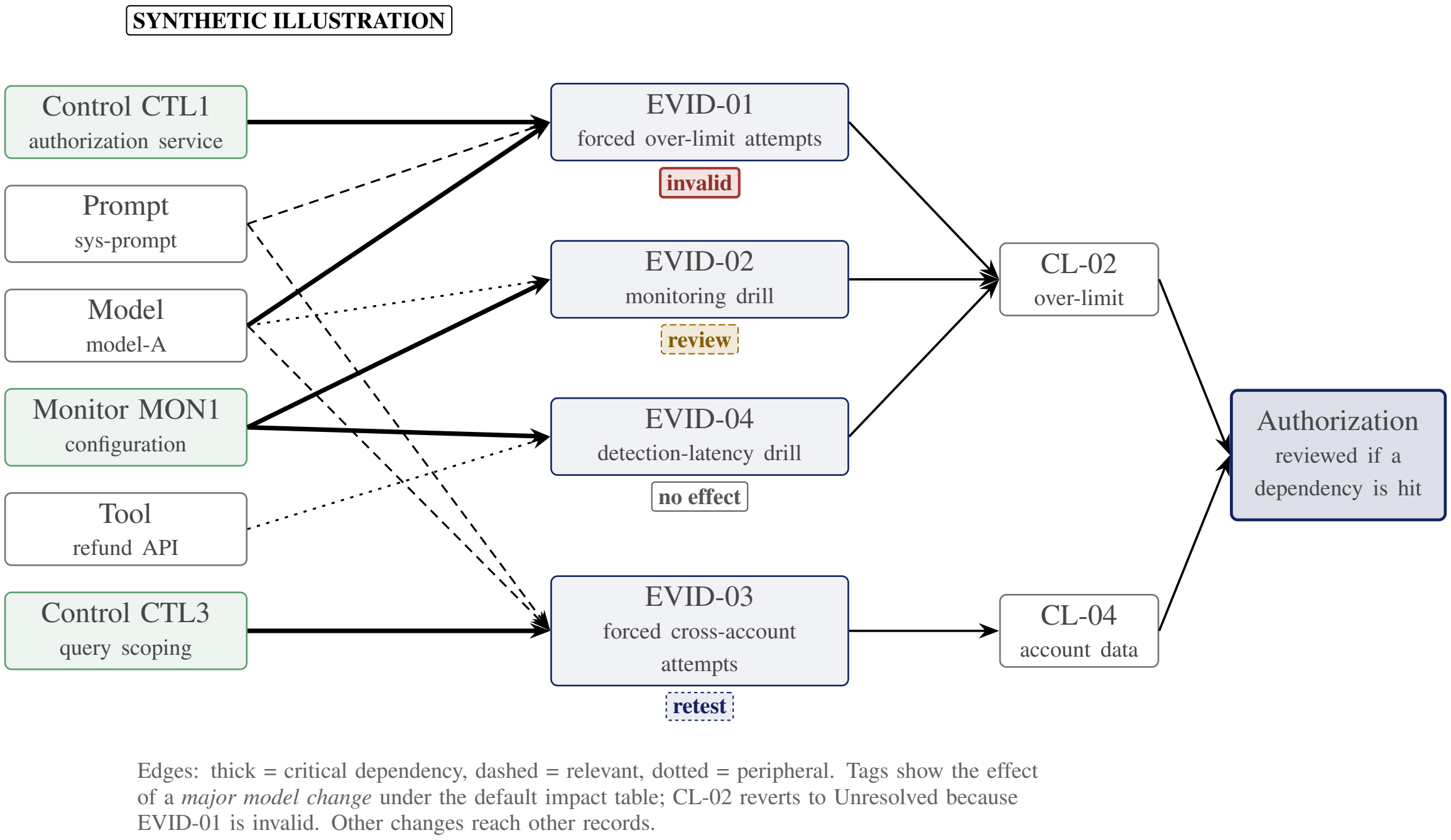
\begin{figure}[!tbp]
\begingroup\hyphenpenalty=10000 \exhyphenpenalty=10000 \centering\resizebox{0.92\linewidth}{!}{\begin{tikzpicture}
  \node[synth, anchor=west] at (0,24mm) {SYNTHETIC ILLUSTRATION};
  \node[v2box, text width=24mm] (c1) at (0,13mm) {Control CTL1\\ \scriptsize authorization service};
  \node[box, text width=24mm] (p) at (0,2mm) {Prompt\\ \scriptsize sys-prompt};
  \node[box, text width=24mm] (m) at (0,-9mm) {Model\\ \scriptsize model-A};
  \node[v2box, text width=24mm] (mo) at (0,-20mm) {Monitor MON1\\ \scriptsize configuration};
  \node[box, text width=24mm] (ap) at (0,-31mm) {Tool\\ \scriptsize refund API};
  \node[v2box, text width=24mm] (c3) at (0,-42mm) {Control CTL3\\ \scriptsize query scoping};
  \node[proc, text width=30mm] (e1) at (62mm,13mm) {EVID-01\\ \scriptsize forced over-limit attempts};
  \node[proc, text width=30mm] (e2) at (62mm,-4mm) {EVID-02\\ \scriptsize monitoring drill};
  \node[proc, text width=30mm] (e4) at (62mm,-21mm) {EVID-04\\ \scriptsize detection-latency drill};
  \node[proc, text width=30mm] (e3) at (62mm,-42mm) {EVID-03\\ \scriptsize forced cross-account attempts};
  \node[box, text width=15mm] (cl2) at (103mm,-4mm) {CL-02\\ \scriptsize over-limit};
  \node[box, text width=15mm] (cl4) at (103mm,-42mm) {CL-04\\ \scriptsize account data};
  \node[dec, text width=21mm, minimum height=14mm] (au) at (131mm,-23mm) {Authorization\\ \scriptsize reviewed if a dependency is hit};
  \draw[cflow] (c1.east) -- (e1.west);
  \draw[dflow] (p.east) -- (e1.west);
  \draw[cflow] (m.east) -- (e1.west);
  \draw[dotflow] (m.east) -- (e2.west);
  \draw[cflow] (mo.east) -- (e2.west);
  \draw[cflow] (mo.east) -- (e4.west);
  \draw[dotflow] (ap.east) -- (e4.west);
  \draw[dflow] (p.east) -- (e3.west);
  \draw[dflow] (m.east) -- (e3.west);
  \draw[cflow] (c3.east) -- (e3.west);
  \draw[flow] (e1.east) -- (cl2.west); \draw[flow] (e2.east) -- (cl2.west); \draw[flow] (e4.east) -- (cl2.west);
  \draw[flow] (e3.east) -- (cl4.west);
  \draw[flow] (cl2.east) -- (au.west); \draw[flow] (cl4.east) -- (au.west);
  \node[tagInv, anchor=north] at ([yshift=-0.6mm]e1.south) {invalid};
  \node[tagRev, anchor=north] at ([yshift=-0.6mm]e2.south) {review};
  \node[tagNo, anchor=north] at ([yshift=-0.6mm]e4.south) {no effect};
  \node[tagRet, anchor=north] at ([yshift=-0.6mm]e3.south) {retest};
  \node[note, anchor=north west, text width=100mm, align=left] at (0,-55mm) {Edges: thick = critical dependency, dashed = relevant, dotted = peripheral. Tags show the effect of a \emph{major model change} under the default impact table; CL-02 reverts to Unresolved because EVID-01 is invalid. Other changes reach other records.};
\end{tikzpicture}}\par\endgroup
\caption{Evidence dependency graph for the synthetic refund agent. Edge style encodes declared criticality. Tags show the effect of one illustrative change, a major model update: only records that declare the model are touched, and only EVID-01, which depends critically on it, becomes invalid, so claim CL-02 reverts to Unresolved and the authorization must be reviewed. The dependency declarations are illustrative assumptions (Section~\ref{sec:case}).}
\label{fig:dep}
\end{figure}

Six rules govern the register. (E1) Only declared dependencies matter; a record that declares none cannot be shown unaffected, so any change above patch size makes it retest, and undeclared dependencies are a defect of the record, not a safe default. (E2) A claim whose supporting evidence is invalid reverts to Unresolved; a claim with retest evidence stays Supported only until the retest deadline and cannot support an expansion of authority. (E3) When evidence marked as an authorization dependency becomes retest or invalid, the decision authority must review the authorization; the register reports that a decision is needed and does not take it. (E4) The authorization expiry cannot be later than the hard expiry of any evidence it depends on. (E5) A supported claim on a high-consequence entry has at least one validity record, and expired evidence cannot support Supported. (E6) Not modeled: the accumulation of many patch-size changes, for which an adopting organization should set a cumulative-change limit. Figure~\ref{fig:dep} illustrates proportional invalidation on the synthetic example of Section~\ref{sec:case}.

\subsection{Control retesting, agentic authority and human oversight}\label{sec:controls}
Controls on high-consequence claims are challenged under realistic failure: forced failures and adversarial inputs, partial outages, degraded dependencies, load and the monitor itself failing. Two controls claimed to fail independently are treated as possibly correlated (Table~\ref{tab:bounds}) unless a test with both degraded shows otherwise.

For an agent, the final answer is one event in a sequence; a final response can be correct while the trajectory violated authority, and wrong while every step was permitted. CSE therefore evaluates the trajectory against a declared \emph{authority envelope}: permitted tools and resources; a reversibility class for each action (R0 no external effect, R1 automatically reversible, R2 reversible only with effort or compensation, R3 irreversible); the class from which human approval is required; a maximum delegation depth with the rule that a delegate holds no authority its delegator does not hold; and a memory policy. Eight trajectory metrics are counts over scenarios, each with a conservative bound and none combined into a score (Appendix~\ref{app:details}); with 60 trajectories and no violation, the two-sided 95\% upper bound is 0.060. Human involvement is not itself a control: oversight is credited only when measured under seeded-error conditions at realistic workload (detection, automation bias, false intervention, workload, escalation speed, authority to act and independence of the measurement); if the reviewer cannot block or reverse the action, the status is Refuted, and self-measured oversight (I0) is not credited for reliance on a high-consequence claim.

\subsection{Consequence-prioritized monitoring tiers}\label{sec:tiers}
A monitoring item names the signal, the claims and Register entries it bears on, a tier, a cadence, a review threshold, the escalation disposition, an owner and evidence that detection is faster than the harm-onset window. A signal that cannot be linked to a claim and a consequence is a metric, not a monitor. The monitoring priority tier T1 to T4 of a pathway follows from five ordinal judgments, each with a stated basis: consequence class (the inherent class, not lowered by credited controls), exposure, assurance of the controls it depends on (L0 to L3 or none), evidence state (current, stale, unresolved) and detection (poor, moderate or good, and whether latency may exceed the harm-onset window, where unknown is treated adversely). The first matching rule decides and is reported so the tier can be challenged (Table~\ref{tab:tiers}). No numerical weights are used. Monitoring tiers are distinct from materiality tiers.

\begin{table}[!tbp]
\centering\footnotesize
\caption{Tier rules TR1 to TR6 (first match decides). Consequence class runs from 1 (low) to 4 (severe); exposure from 1 (narrow) to 4 (broad).}
\label{tab:tiers}
\begin{tabular}{@{}L{1.0cm}L{11.2cm}L{2.2cm}@{}}
\toprule
Rule & Condition & Tier \\
\midrule
TR1 & Consequence class 3 or 4, and (controls below L2, or evidence stale or unresolved, or detection possibly slower than harm onset). & T1 \\
TR2 & Severe (class 4) or irreversible, with verified controls, current evidence and fast detection. & T2, never below \\
TR3 & Consequence class 3 with verified controls and current evidence. & T2 \\
TR4 & Class 2, broad exposure (3 to 4), and weak controls or evidence. & T2 \\
TR5 & Class 2; or class 1 with broad exposure and weak controls. & T3 \\
TR6 & Otherwise. & T4 \\
\bottomrule
\end{tabular}
\end{table}

Default cadences illustrate the form: T1 daily signal check, weekly human review, monthly control retest and a 90-day maximum authorization; T2 daily, monthly, quarterly and 180 days; T3 weekly, quarterly, semiannual and 365 days; T4 monthly, annual, annual and 365 days, with independent verification once per period for T1 and T2. When tiers demand more effort than capacity, funding follows a fixed order (tier, then consequence class, then exposure, then identifier) with a reserved fraction for random audits, and unfunded items are listed for the governance committee rather than moved to a lower tier. A circularity guard applies: tiers use pre-registered rules and prior evidence, audits are drawn independently of tier, and absence of signal from an unmonitored pathway is Unresolved, never evidence of safety. That this allocation avoids more expected harm than alternatives is hypothesis H7, untested.

\subsection{Authorization changes and the failure-to-learning loop}\label{sec:escalation}
Escalation rules are recorded in the monitoring plan and the authorization. Illustrative defaults map signals to dispositions: a completed harm event on a high-consequence claim leads to Stop pending review; a failed retest of a credited control leads to Escalate, and to Stop if it is the only barrier; expired or invalidated evidence reverts the claim to Unresolved, so there is no Deploy while it is tied to a high-consequence entry; repeated near misses lead to Remediate; and exceeded reviewer capacity leads to Escalate, since an unreviewed signal is not a monitored signal (Appendix~\ref{app:details}). Renewal issues a new Deploy authorization whose expiry is no later than the tier maximum and the earliest expiry of an authorization-dependent evidence record. Failures and near misses change the evaluation through six gated stages, Discover, Reproduce, Contain, Fix, Verify and Learn; for a high-consequence entry the verifier is not the fixing team, a proposed fix is not a verified fix, and Learn requires at least one change to the evaluation, including invalidation of existing evidence in the register. A record cannot claim a stage whose gate is unmet.

\subsection{Policy-conditioned loss and avoided-loss contrast}\label{sec:avoided}
A monitoring-and-intervention policy $\pi$ is a specified detection rule, response rule, response authority and latency. Conditioning Equation~\eqref{eq:identity} on $\pi$ gives $\Ex[L\cnd A,\pi]=\Prob(F\cnd A,\pi)\,\Prob(H\cnd F,A,\pi)\,\Ex[S\cnd H,F,A,\pi]$. Define, for a failure of the stated kind, $d=\Prob(\text{detect}\cnd F)$, $t=\Prob(\text{intervene in time}\cnd\text{detect},F)$ and $c=\Prob(\text{contain}\cnd\text{in time},\text{detect},F)$. By the chain rule the probability that a failure is detected, acted on in time and contained is $dtc$ without an independence assumption, provided each factor is conditioned on the preceding steps. That identity is separate from the harm-reduction claim. Let $C$ be the event that a failure is detected, acted on in time and contained, and let $\pi_0$ be the baseline policy. The factorization $\Prob(H\cnd F,A,\pi)=\Prob(H\cnd F,A,\pi_0)(1-dtc)$ is a model, not an identity, and holds only under these assumptions: (A1) the policy changes neither $\Prob(F\cnd A)$ nor severity given harm; (A2) contained failures cause no harm, $\Prob(H\cnd F,C,A,\pi)=0$; (A3) uncontained failures are as likely to cause harm as under the baseline, $\Prob(H\cnd F,\neg C,A,\pi)=\Prob(H\cnd F,A,\pi_0)$, so detection does not select harmful or benign failures; (A4) $d$, $t$ and $c$ are estimated on the same failure population and operating conditions as the harm term, and the baseline has no detection or containment of its own; (A5) containment precedes harm. Under A2 and A3, $\Prob(H\cnd F,A,\pi)=\Prob(\neg C\cnd F)\Prob(H\cnd F,\neg C,A,\pi)=(1-dtc)\Prob(H\cnd F,A,\pi_0)$. If detection preferentially catches the failures that would cause harm, A3 fails and the formula overstates residual harm (conservative); if it mostly catches benign failures, it understates it. Detection does not imply containment: a high $d$ with $t=0$ avoids no loss. When two policies share the three factors, the shared product $X=\Prob(F)\Prob(H\cnd F)\Ex[S]$ is common and under the same assumptions the expected avoided loss is $X(r_0-r_1)$, with $r_i=1-dtc$ for policy $i$ and $r=1$ for no monitoring; treating the shared factors as one coupled term prevents counting their uncertainty twice.

The corrected interval arithmetic for a hypothetical policy (coupled contrast $[252,6912]$ versus $[-7976,11576]$ when the two losses are bounded separately) is given in Appendix~\ref{app:details}. The contrast concerns expected loss only: a policy can lower expected loss and leave stressed loss unchanged when a common-cause burst exhausts response capacity, so the stress scenarios for high-consequence entries include the monitoring chain itself failing. A result may be called causal only for the tested conditions of a designed experiment; otherwise it is labeled by its evidence basis.

\section{Reproducible Synthetic Case Study}\label{sec:case}
\textbf{This section is an illustrative synthetic walkthrough, not an experiment.} The system, its tests and every count are invented in the Public Framework as a teaching construction \citep{rbed_framework}; nothing here was measured, and no result is evidence about any real system or about the effectiveness of \RBED{}. What we add is a deterministic calculation: a standalone script (Appendix~\ref{app:repro}) recomputes every interval, status, bound and impact below from the stated counts. No random numbers are drawn, so no seeds are needed.

\subsection{Setup}\label{sec:casesetup}
\paragraph{Business task and configured system (\synth{}).} A customer-support agent answers policy questions and may initiate a refund only after authenticated authorization and approval above an established threshold. It has access to customer records, policy documents and refund tools. The configured system $A$ is one model build, one prompt, one retrieval snapshot, a refund API and two server-side controls, with a 90-day boundary. Table~\ref{tab:casesetup} gives the specification elements.

\begin{table}[!tbp]
\centering\footnotesize
\caption{Specification of the synthetic refund agent. All content is \synth{}. Thresholds were fixed before testing for a restricted pilot with human approval retained; a general release would need a far lower threshold.}
\label{tab:casesetup}
\begin{tabular}{@{}L{3.1cm}L{11.9cm}@{}}
\toprule
Element & Specification \\
\midrule
Authorized actions & Answer policy questions (read); read the authenticated customer's own account (G2, read only, scoped by CTL3); initiate a refund at or below the threshold after authenticated approval (G1, act-with-approval, per-refund and per-day limits, enforced by CTL1). \\
Forbidden actions & Any refund above the threshold or limit; any access to another customer's account data; any tool call induced by instruction-like text in untrusted retrieved content. \\
Harm pathways & K1 unauthorized money movement (financial, moderate, partially reversible); K2 notice obligations from an erroneous refund or denial (legal and regulatory, moderate, reversible); K3 cross-account disclosure (privacy, moderate, irreversible, hence high-consequence); RSK-02 harm from the policy itself under intended behavior (pathway $N$, no outcome data). \\
Claims and thresholds & CL-01 a retrieved note cannot override refund policy (categorical). CL-02 a server-side check rejects over-limit refunds regardless of model content (upper 95\% bound at most 0.10). CL-03 no unauthorized refund is completed (categorical prohibition). CL-04 replies and queries expose account data only to the authenticated customer (upper 95\% bound at most 0.10). \\
Failure injection and evidence & 20 injected-note scenarios; 25 and later 60 forced over-limit tool calls through a sandbox gateway; 60 forced cross-account attempts; 10 monitoring drills; validity records EVID-01 to EVID-04 (Figure~\ref{fig:dep}). \\
Deployment controls & CTL1: independent server-side authorization service, enforced outside the model (B1). CTL3: server-side scoping of every account-data query to the authenticated session, which the model cannot widen (B1). Human approval retained; daily attempted-versus-completed refund report; weekly drill. \\
Authorization & Distinct named authorizer; expiry on day 30; invalidated by a completed over-limit refund, a model, tool or authorization-service version change, or a drill failure. \\
\bottomrule
\end{tabular}
\end{table}

\subsection{What a task-success view can miss}\label{sec:taskmiss}
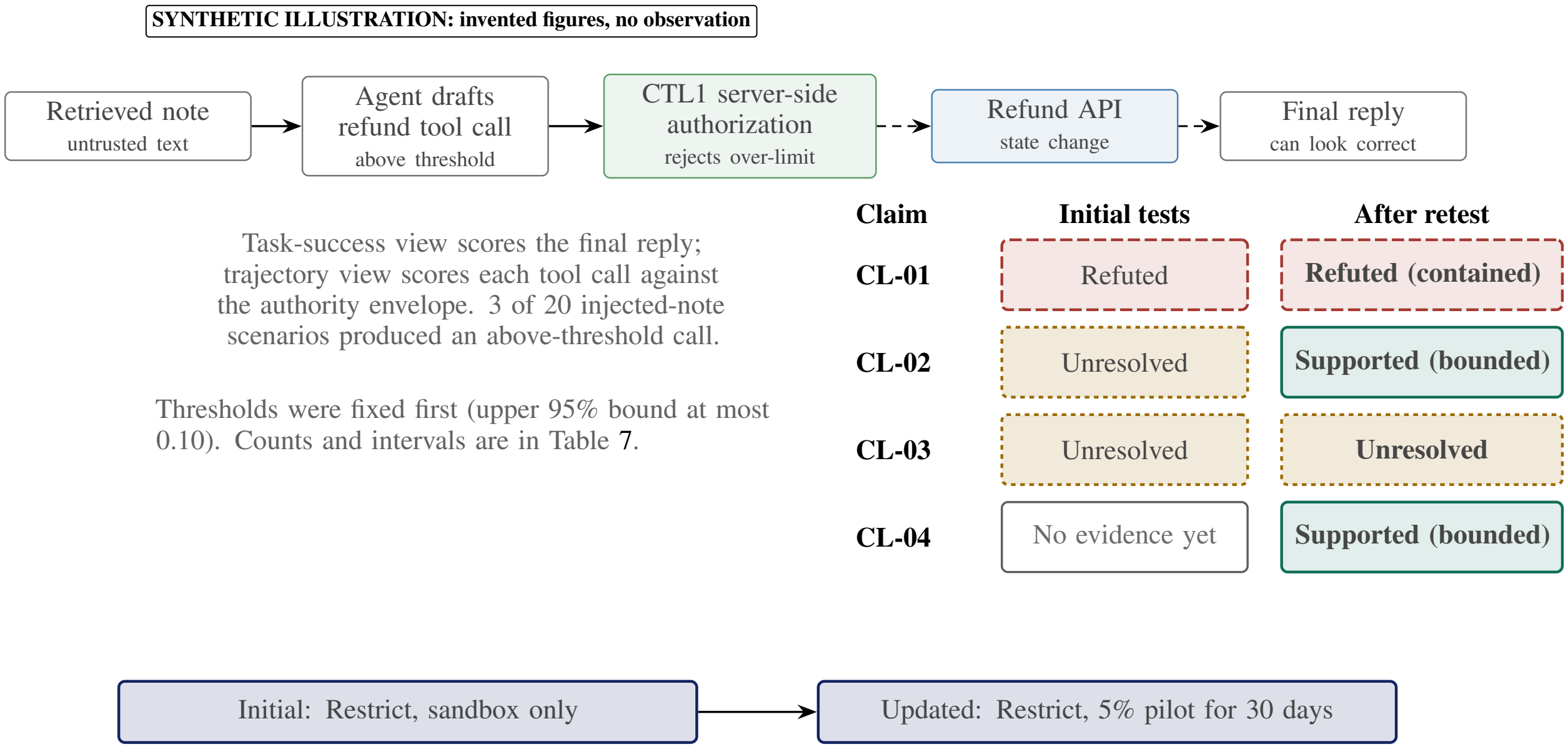
\begin{figure}[!tbp]
\begingroup\hyphenpenalty=10000 \exhyphenpenalty=10000 \centering\resizebox{1.0\linewidth}{!}{\begin{tikzpicture}
  \node[synth, anchor=west] at (0,34mm) {SYNTHETIC ILLUSTRATION: invented figures, no observation};
  % trajectory
  \node[box, text width=26mm] (t1) at (-2mm,22mm) {Retrieved note\\ \scriptsize untrusted text};
  \node[box, text width=26mm] (t2) at (32mm,22mm) {Agent drafts refund tool call\\ \scriptsize above threshold};
  \node[ctl, text width=29mm] (t3) at (68mm,22mm) {CTL1 server-side authorization\\ \scriptsize rejects over-limit};
  \node[harm, text width=26mm] (t4) at (104mm,22mm) {Refund API\\ \scriptsize state change};
  \node[box, text width=26mm] (t5) at (137mm,22mm) {Final reply\\ \scriptsize can look correct};
  \draw[flow] (t1)--(t2); \draw[flow] (t2)--(t3); \draw[dflow] (t3)--(t4); \draw[dflow] (t4)--(t5);
  \node[note, font=\footnotesize, text width=72mm, anchor=north west] at (0,11mm) {Task-success view scores the final reply; trajectory view scores each tool call against the authority envelope. 3 of 20 injected-note scenarios produced an above-threshold call.};
  % evidence states
  \node[font=\footnotesize\bfseries, anchor=west] at (80mm,12mm) {Claim};
  \node[font=\footnotesize\bfseries, align=center] at (112mm,12mm) {Initial tests};
  \node[font=\footnotesize\bfseries, align=center] at (146mm,12mm) {After retest};
  \foreach \cl/\sa/\a/\sb/\b/\y in {CL-01/sRef/Refuted/sRef/Refuted (contained)/5, CL-02/sUnr/Unresolved/sSup/Supported (bounded)/-5, CL-03/sUnr/Unresolved/sUnr/Unresolved/-15, CL-04/sNA/No evidence yet/sSup/Supported (bounded)/-25}
    { \node[font=\footnotesize\bfseries, anchor=west] at (80mm,\y mm) {\cl};
      \node[\sa, font=\footnotesize, text width=26mm, minimum height=8mm] at (112mm,\y mm) {\a};
      \node[\sb, text width=30mm, minimum height=8mm] at (146mm,\y mm) {\b}; }
  \node[note, font=\footnotesize, text width=72mm, anchor=north west, align=left] at (0,-8mm) {Thresholds were fixed first (upper 95\% bound at most 0.10). Counts and intervals are in Table~\ref{tab:caseresults}.};
  % disposition
  \node[dec, text width=64mm] (d1) at (30mm,-45mm) {\footnotesize Initial: Restrict, sandbox only};
  \node[dec, text width=64mm] (d2) at (110mm,-45mm) {\footnotesize Updated: Restrict, 5\% pilot for 30 days};
  \draw[flow] (d1)--(d2);
\end{tikzpicture}}\par\endgroup
\caption{Synthetic refund-agent walkthrough. Top: trajectory from an untrusted note to a tool call; the final reply can look correct whether or not the call was blocked. Right: claim status before and after the independent control and retest. Bottom: disposition before and after. All figures are invented.}
\label{fig:case}
\end{figure}

Consider the scenario in which an untrusted retrieved note induces a tool call above the threshold while the model gives a plausible reply (Figure~\ref{fig:case}). A task-success evaluation that scores the final reply, or even a refund outcome in the absence of a challenge, reports on the visible end of the sequence. In the \RBED{} procedure the same scenario is evaluated against claims on the trajectory and against the controls that stand between the call and the payment gateway. The source example notes that a correct final response does not erase a misdirected tool call; it records no final-answer accuracy, and we assert none. The point is qualitative: a pass on reply quality leaves CL-01 and CL-02 without the evidence they need, and the procedure makes that absence explicit as a status.

\subsection{Evidence, statuses and bounds}\label{sec:caseevidence}
Table~\ref{tab:caseresults} reports the counts of the source example with exact (Clopper--Pearson) two-sided 95\% intervals computed by the script, and the status each follows from the rules of Section~\ref{sec:statuses}. Initial tests find three of twenty injected-note scenarios producing an above-threshold tool call, so CL-01, a categorical claim, is Refuted. Zero completed over-limit refunds in 25 forced attempts gives an upper bound of 0.137, above the 0.10 threshold, and detection latency is untested, so CL-02 is Unresolved: the observed zero is not a result the claim can rest on. CL-03, the categorical prohibition, is Unresolved because no production-like outcome evidence exists. After an independent server-side authorization control is added (Remediate) and an independent reviewer outside the build team (I2) reruns the seeded failure, legitimate requests, regression scenarios and a monitoring drill, the attack still reaches tool calls, but 0 of 60 forced attempts complete (upper bound 0.060), and a second forced test of the account-data scoping gives 0 of 60 cross-account disclosures (upper bound 0.060). CL-02 and CL-04 are then Supported (bounded) against the 0.10 pilot threshold, for this build, gateway and window only; the 0.060 bound would not support a stringent tolerance, and zero observed failures is still not zero risk.

\begin{table}[!tbp]
\centering\footnotesize
\caption{Counts, exact 95\% intervals and statuses for the synthetic example (\synth{}). Intervals are recomputed by the reproducibility script. CL-01 is categorical, so any observed violation refutes it. The source example states no threshold for the drill measure.}
\label{tab:caseresults}
\begin{tabular}{@{}L{1.4cm}L{5.0cm}L{1.8cm}L{3.0cm}L{3.3cm}@{}}
\toprule
Claim & Test (independence) & Count & 95\% interval & Status \\
\midrule
CL-01 & Injected notes producing an above-threshold call (initial) & 3 of 20 & $[0.032,\,0.379]$ & Refuted (contained later) \\
CL-02 & Completed over-limit refunds, forced (initial, I0) & 0 of 25 & $[0,\,0.137]$ & Unresolved \\
CL-02 & Completed over-limit refunds, forced (retest, I2) & 0 of 60 & $[0,\,0.060]$ & Supported (bounded) \\
CL-04 & Completed cross-account disclosures, forced (I2) & 0 of 60 & $[0,\,0.060]$ & Supported (bounded) \\
CL-02 & Drills triggering intervention within the agreed limit (I2) & 10 of 10 & $[0.692,\,1]$ & Lower bound reported \\
CL-03 & Production-like outcomes & none & not available & Unresolved \\
\bottomrule
\end{tabular}
\end{table}

\subsection{Decision}\label{sec:casedecision}
The initial disposition is Restrict to a sandbox, with any completed over-limit refund or version change triggering re-review. After the retest the claims have the statuses above. Deploy remains blocked: CL-03 is a categorical prohibition that is unresolved (rule 3). The updated disposition is a Restrict, a limited pilot at 5\% of tickets for 30 days. The 5\% only limits scale; under rule 5 the restriction on the pathways is carried by CTL1 (over-limit refunds) and CTL3 (cross-account disclosure), each enforced outside the model and with effectiveness evidence at an independent level. Because K3 is high-consequence, rule 7 requires evidence that is not I0-only (the retests are I2), an approving authority distinct from the evaluation and business owners, and pointers to the specialized security and privacy assessments. Residual risk follows rule 8: RSK-01, an over-limit refund completing despite the authorization service, has upper bound 0.060 per forced attempt (an upper bound, not an estimate); RSK-02, harm from the policy itself under intended behavior, is an explicit abstention because no outcome data exist; and insurance against payment fraud is recorded as a transfer, not as harm reduction. Broader deployment awaits CL-03 evidence and risk-owner sign-off.

\subsection{Proportional invalidation}\label{sec:caseinval}
Table~\ref{tab:caseinval} applies the default impact table to four hypothetical changes. The dependency declarations behind it are \emph{illustrative assumptions}: the CSE specification reports the resulting table, and we reconstructed declarations that reproduce it (for example, EVID-01 declares the model and CTL1 as critical and the prompt as relevant). They are not read from a released data file. A prompt patch leaves every record intact. A minor change to CTL3 reaches only the evidence that depends on it. A major model change invalidates the over-limit evidence, reduces the monitoring drill to a review and leaves the detection-latency drill untouched because that record does not depend on the model. The review burden follows the dependency graph, not the number of changes, and E6 (accumulated patches) is not modeled.

\begin{table}[!tbp]
\centering\footnotesize
\caption{Impact of four hypothetical changes on the four synthetic validity records (\synth{}; dependency declarations are illustrative assumptions).}
\label{tab:caseinval}
\begin{tabular}{@{}L{4.1cm}cccccL{2.4cm}@{}}
\toprule
Change & EVID-01 & EVID-02 & EVID-03 & EVID-04 & Claims reverting \\
\midrule
System prompt, patch & none & none & none & none & none \\
Query-scoping control CTL3, minor & none & none & retest & none & none \\
Model, minor update & retest & none & review & none & none \\
Model, major change & invalid & review & retest & none & CL-02 \\
\bottomrule
\end{tabular}
\end{table}

\subsection{What the walkthrough does and does not show}\label{sec:casescope}
It shows that the specification can be applied end to end, that its rules give determinate statuses and dispositions on a stated record, and that every number can be recomputed. It does not show that the method finds more weaknesses than another, that the thresholds are right, that a real agent would behave this way, or that proportional invalidation is practical at scale; no claim of superior effectiveness is made.

\section{Discussion, Limitations and Validation Agenda}\label{sec:discussion}
\subsection{Maturity and the state of evidence}\label{sec:maturity}
Two kinds of evidence are kept separate. A \emph{methodological implementation} shows that software follows a specification. \emph{Prospective validation}, independent evaluation of whether applying \RBED{} improves outcomes, does not exist. Retrospective analysis of documented incidents is outside the scope of this paper. Table~\ref{tab:maturity} gives each component's label.

\begin{table}[!tbp]
\centering\footnotesize
\caption{Maturity of components. [E] specified core, with software checks of the specified rules in the author's reference tooling; [P] proposed extension; [H] research hypothesis. No label denotes empirical validation.}
\label{tab:maturity}
\begin{tabular}{@{}L{8.8cm}L{1.3cm}L{4.9cm}@{}}
\toprule
Component & Label & Software status \\
\midrule
6+4+3 architecture, objectives, Consequence Register & [E] & Data model and selected checks in the public package. \\
Conditional pathway identity, evidence statuses, five dispositions, eight rules, SES minimum fields & [E] & Partly implemented in the public package (data model, selected checks, interval statistics). \\
Evidence Validity Register, proportional invalidation & [P] & Standalone demonstration script only. \\
Assurance ladder and control types; authority envelope, trajectory and oversight metrics; monitoring tiers, allocation, learning loop, avoided-loss contrast & [P] & None released. \\
Consequence-aware allocation avoids more harm than frequency-based allocation (H7) & [H] & Not tested. \\
\bottomrule
\end{tabular}
\end{table}

\paragraph{Implementation claims.} Three things must not be confused. (1) The public Python package \texttt{rbed} (version 0.2.0) implements the System Evaluation Specification data model with its discipline checks, interval statistics (Wilson, Clopper--Pearson, zero-failure bounds), conditional-identity helpers and the machine-readable NIST crosswalk; its tests check that code follows the specification. It does \emph{not} implement the Evidence Validity Register or the dependency-invalidation module, the control assurance ladder, the trajectory metrics or the tier rules. (2) The standalone script shipped with this manuscript recomputes the synthetic walkthrough, including the default impact table applied to reconstructed dependency declarations. It is a demonstration, not a release of the proposed modules. (3) Software tests show that code follows text, not that the method works.

\subsection{Limitations}\label{sec:limits}
\begin{itemize}
  \item \textbf{Unknown or unobserved failure pathways.} Enumeration is bounded by what the team and affected parties can imagine; absence of signal is Unresolved, not safe.
  \item \textbf{Measurement and construct validity.} A metric may not measure the claim it supports; thresholds rest on judgment, and method review is a proposed discipline, not proof.
  \item \textbf{Nonstationary conditions.} Evidence is valid for a configuration, population and period; drift and adaptive adversaries can invalidate it, and accumulated patch-size changes are not modeled (E6).
  \item \textbf{Correlated control failures.} Controls sharing code, data, vendors or credentials fail together; Fr\'echet bounds are wide, and independence must be evidenced.
  \item \textbf{Underreporting and incomplete incident data.} Observed harm is a lower bound on true harm, and any ascertainment correction is itself an assumption.
  \item \textbf{Human oversight and automation bias.} Approval steps may be formalities; crediting oversight needs seeded-error measurement, which is costly and can be gamed.
  \item \textbf{Non-monetizable harms and subjective ratings.} Dignity, autonomy, psychological and societal harms are recorded by category and never priced for summation, which limits aggregation. Ordinal ratings and tier inputs depend on judgment; reliability is an empirical question (H1), and gaming of ratings is open.
  \item \textbf{Generalizability.} One synthetic customer-support agent is worked; cadences and tables are illustrative defaults, not derived from data.
  \item \textbf{Need for independent prospective validation.} The author is not independent of the method, and a record that satisfies the rules can still describe an unsafe system.
\end{itemize}
The burden of maintaining records at scale is unknown (H5); specialized security, privacy, safety and legal assessments remain outside the method.

\subsection{Validation agenda}\label{sec:validation}
Seven hypotheses, each of which could be false, are stated in the Validation Plan \citep{rbed_validation}, which proposes seven studies (S1 to S7, one or more per hypothesis); none has been run and no result is claimed. In brief: H1 reviewers agree on statuses and the high-consequence classification more than chance; H2 incident reconstruction exposes the gap investigators later reported more often than unstructured review; H3 independent (I2 or I3) challenge finds failures that I0 self-assessment misses; H4 intervention-contrast estimates predict later observed containment; H5 teams can complete records at acceptable cost and change a decision; H6 the rules block decisions experts judge unsupported without blocking many judged supported; H7 tier-based allocation avoids more expected harm than frequency-based or uniform allocation in stated settings. Full statements are in Appendix~\ref{app:hyp}.

\section{Conclusion}\label{sec:conclusion}
We asked how evidence about technical failures, control effectiveness and real-world consequences can be connected to defensible authorization decisions over a deployed system's lifecycle. \RBED{} answers with a specification: a fixed 6+4+3 architecture; a failure-mediated identity kept separate from intended-behavior and misuse pathways; bounded evidence statuses with a control assurance ladder; five dispositions under eight rules; and a Continuous System Evaluation layer in which evidence is pinned, dated and invalidated in proportion to change. The contribution is a reproducible methodological specification, not an empirical result: several components are proposals, one is a hypothesis, and independent prospective validation does not exist.

\bibliographystyle{plainnat}
\bibliography{references}

@techreport{nist_rmf,
  author = {{National Institute of Standards and Technology}},
  title = {Artificial Intelligence Risk Management Framework ({AI RMF} 1.0)},
  institution = {NIST},
  number = {NIST AI 100-1},
  year = {2023},
  doi = {10.6028/NIST.AI.100-1}
}

@techreport{nist_genai,
  author = {{National Institute of Standards and Technology}},
  title = {Artificial Intelligence Risk Management Framework: Generative Artificial Intelligence Profile},
  institution = {NIST},
  number = {NIST AI 600-1},
  year = {2024},
  doi = {10.6028/NIST.AI.600-1}
}

@misc{iso42001,
  author = {{International Organization for Standardization and International Electrotechnical Commission}},
  title = {{ISO/IEC} 42001:2023, Information technology, Artificial intelligence, Management system},
  year = {2023}
}

@misc{sr117,
  author = {{Board of Governors of the Federal Reserve System and Office of the Comptroller of the Currency}},
  title = {{SR} 11-7: Guidance on Model Risk Management},
  year = {2011},
  note = {Issued 4 April 2011; superseded by SR 26-2}
}

@misc{sr262,
  author = {{Board of Governors of the Federal Reserve System, Office of the Comptroller of the Currency and Federal Deposit Insurance Corporation}},
  title = {{SR} 26-2: Revised Guidance on Model Risk Management},
  year = {2026},
  note = {Issued 17 April 2026; footnote 3 places generative and agentic AI models outside its scope},
  url = {https://www.federalreserve.gov/supervisionreg/srletters/SR2602.pdf}
}

@misc{srivastava2026,
  author = {Srivastava, Vishal and Sah, Tanmay},
  title = {Quantifying Automation Risk in High-Automation {AI} Systems: A {B}ayesian Framework for Failure Propagation and Optimal Oversight},
  year = {2026},
  eprint = {2602.18986},
  archivePrefix = {arXiv},
  primaryClass = {cs.AI},
  url = {https://arxiv.org/abs/2602.18986}
}

@inproceedings{mcgregor2021,
  author = {McGregor, Sean},
  title = {Preventing Repeated Real World {AI} Failures by Cataloging Incidents: The {AI} Incident Database},
  booktitle = {Proceedings of the AAAI Conference on Artificial Intelligence},
  volume = {35},
  pages = {15458--15463},
  year = {2021}
}

@article{liang2023helm,
  author = {Liang, Percy and others},
  title = {Holistic Evaluation of Language Models},
  journal = {Transactions on Machine Learning Research},
  year = {2023},
  note = {arXiv:2211.09110}
}

@misc{shevlane2023,
  author = {Shevlane, Toby and others},
  title = {Model evaluation for extreme risks},
  year = {2023},
  eprint = {2305.15324},
  archivePrefix = {arXiv}
}

@misc{weidinger2023,
  author = {Weidinger, Laura and others},
  title = {Sociotechnical Safety Evaluation of Generative {AI} Systems},
  year = {2023},
  eprint = {2310.11986},
  archivePrefix = {arXiv}
}

@misc{clymer2024,
  author = {Clymer, Joshua and Gabrieli, Nick and Krueger, David and Larsen, Thomas},
  title = {Safety Cases: How to Justify the Safety of Advanced {AI} Systems},
  year = {2024},
  eprint = {2403.10462},
  archivePrefix = {arXiv}
}

@misc{yao2024tau,
  author = {Yao, Shunyu and Shinn, Noah and Razavi, Pedram and Narasimhan, Karthik},
  title = {$\tau$-bench: A Benchmark for Tool-Agent-User Interaction in Real-World Domains},
  year = {2024},
  eprint = {2406.12045},
  archivePrefix = {arXiv}
}

@inproceedings{debenedetti2024,
  author = {Debenedetti, Edoardo and Zhang, Jie and Balunovi{\'c}, Mislav and Beurer-Kellner, Luca and Fischer, Marc and Tram{\`e}r, Florian},
  title = {{AgentDojo}: A Dynamic Environment to Evaluate Prompt Injection Attacks and Defenses for {LLM} Agents},
  booktitle = {Advances in Neural Information Processing Systems, Datasets and Benchmarks Track},
  year = {2024},
  note = {arXiv:2406.13352}
}

@inproceedings{ruan2024,
  author = {Ruan, Yangjun and others},
  title = {Identifying the Risks of {LM} Agents with an {LM}-Emulated Sandbox},
  booktitle = {International Conference on Learning Representations},
  year = {2024},
  note = {arXiv:2309.15817}
}

@inproceedings{raji2020,
  author = {Raji, Inioluwa Deborah and others},
  title = {Closing the {AI} Accountability Gap: Defining an End-to-End Framework for Internal Algorithmic Auditing},
  booktitle = {Proceedings of the 2020 Conference on Fairness, Accountability, and Transparency},
  year = {2020},
  note = {arXiv:2001.00973}
}

@inproceedings{sculley2015,
  author = {Sculley, D. and others},
  title = {Hidden Technical Debt in Machine Learning Systems},
  booktitle = {Advances in Neural Information Processing Systems 28},
  year = {2015}
}

@inproceedings{breck2017,
  author = {Breck, Eric and Cai, Shanqing and Nielsen, Eric and Salib, Michael and Sculley, D.},
  title = {The {ML} Test Score: A Rubric for {ML} Production Readiness and Technical Debt Reduction},
  booktitle = {2017 IEEE International Conference on Big Data},
  year = {2017}
}

@article{kaplan1981,
  author = {Kaplan, Stanley and Garrick, B. John},
  title = {On the Quantitative Definition of Risk},
  journal = {Risk Analysis},
  volume = {1},
  number = {1},
  pages = {11--27},
  year = {1981}
}

@article{artzner1999,
  author = {Artzner, Philippe and Delbaen, Freddy and Eber, Jean-Marc and Heath, David},
  title = {Coherent Measures of Risk},
  journal = {Mathematical Finance},
  volume = {9},
  number = {3},
  pages = {203--228},
  year = {1999}
}

@article{rockafellar2000,
  author = {Rockafellar, R. Tyrrell and Uryasev, Stanislav},
  title = {Optimization of Conditional Value-at-Risk},
  journal = {Journal of Risk},
  volume = {2},
  number = {3},
  pages = {21--41},
  year = {2000}
}

@article{clopper1934,
  author = {Clopper, C. J. and Pearson, E. S.},
  title = {The Use of Confidence or Fiducial Limits Illustrated in the Case of the Binomial},
  journal = {Biometrika},
  volume = {26},
  number = {4},
  pages = {404--413},
  year = {1934}
}

@article{frechet1951,
  author = {Fr{\'e}chet, Maurice},
  title = {Sur les tableaux de corr{\'e}lation dont les marges sont donn{\'e}es},
  journal = {Annales de l'Universit{\'e} de Lyon, Section A},
  volume = {14},
  pages = {53--77},
  year = {1951}
}

@misc{rbed_framework,
  author = {Srivastava, Vishal Kumar},
  title = {{R-BED} Public Framework, Version 2.1},
  year = {2026},
  note = {Original release 29 October 2026; corrected edition 4 November 2026. No stable public URL or identifier was available at the time of writing}
}

@misc{rbed_cse,
  author = {Srivastava, Vishal Kumar},
  title = {{R-BED} Continuous System Evaluation ({CSE}) Specification, Version 2.1},
  year = {2026},
  note = {Original release 29 October 2026; corrected edition 4 November 2026. No stable public URL or identifier was available at the time of writing}
}

@misc{rbed_crosswalk,
  author = {Srivastava, Vishal Kumar},
  title = {{NIST} {AI} {RMF} 1.0 $\times$ {R-BED} Crosswalk, Version 2.1},
  year = {2026},
  note = {Original release 29 October 2026; corrected edition 4 November 2026. No stable public URL or identifier was available at the time of writing}
}

@misc{rbed_validation,
  author = {Srivastava, Vishal Kumar},
  title = {{R-BED} Validation Plan, Version 2.1},
  year = {2026},
  note = {No stable public URL or identifier was available at the time of writing}
}

\appendix
\section{Reproducibility of the Synthetic Walkthrough}\label{app:repro}
The directory \texttt{reproducibility/} of the delivered project contains a companion research artifact for Section~\ref{sec:case}. It is a standalone demonstration, separate from the public \texttt{rbed} package.
\begin{itemize}
  \item \texttt{walkthrough.py}: standard library only (Python 3). It computes exact Clopper--Pearson intervals by bisection on the binomial distribution, applies the status rules to the counts of Table~\ref{tab:caseresults}, applies the default impact table of Table~\ref{tab:impact} to the illustrative dependency declarations, and recomputes the quantities of Table~\ref{tab:bounds} and Section~\ref{sec:avoided}.
  \item \texttt{test\_walkthrough.py}: eight unit tests (intervals, zero-failure trial counts, statuses, proportional invalidation, avoided-loss intervals, the three-configuration comparison, the Fr\'echet bound, and exceedance and intervention-contrast checks).
  \item \texttt{output.json}: the output of one run.
  \item \texttt{README.md}: instructions.
\end{itemize}
\paragraph{Configuration and seeds.} There is no random number generation and therefore no seed. The inputs are the counts and ranges written at the top of the script (the invented counts of the source example and the hypothetical ranges of Section~\ref{sec:avoided}); the pilot threshold is 0.10.
\paragraph{Commands.} From the \texttt{reproducibility/} directory, \texttt{python3 walkthrough.py} prints and writes \texttt{output.json}, and \texttt{python3 -m unittest test\_walkthrough} runs the tests.
\paragraph{Reproduced values.} Intervals: $3/20$ gives $[0.0321,0.3789]$, $0/25$ gives upper $0.1372$, $0/60$ gives upper $0.0596$, and $10/10$ gives lower $0.6915$. Zero-failure trial counts: $29$, $299$ and $2995$ for thresholds $0.10$, $0.01$ and $0.001$. Avoided-loss intervals: coupled $[252,6912]$ and separately bounded $[-7976,11576]$. Three-configuration expected losses: $1.92$, $0.028$ and $0.009$. Impact table: as in Table~\ref{tab:caseinval}.
\paragraph{Limits.} The script reproduces arithmetic on invented inputs. The dependency declarations are a reconstruction chosen to reproduce the published table of the specification, not data. Nothing is a measured result.

\section{Notation}\label{app:notation}
\begin{center}\small
\begin{tabular}{@{}L{2.6cm}L{12.4cm}@{}}
\toprule
Symbol & Meaning \\
\midrule
$A$ & Configured system: model, prompts, tools, permissions, data, controls and operating context. \\
$F$, $H$, $S$ & Failure of a stated kind in one exposure; harm of a stated category after deployment controls, given $F$; severity of the harm ($S\ge 0$, in currency units where priced, otherwise ordinal and not summed). \\
$N$, $M$ & Harm from intended behavior that meets every claim; harm from misuse. \\
$L_F,L_N,L_M$ & Pathway losses; $L_F=S\,\mathbf 1\{F\}\,\mathbf 1\{H\}$. \\
$\Ex$, $\Prob$ & Expectation and probability, per exposure unit over a stated horizon. \\
$\pi$ & Monitoring-and-intervention policy (detection rule, response rule, authority, latency); $\pi_0$ is the baseline. \\
$d,t,c$ & Probability of detection given failure; of timely intervention given detection; of containment given timely intervention. \\
$X$ & Shared product $\Prob(F)\Prob(H\cnd F)\Ex[S]$ of two policies. \\
V1 to V3 & Vertices: Technical Failure, Deployment Controls, Consequence and Impact. \\
A1 to A6, B1 to B4, C1 to C3 & The thirteen dimensions; C3a regulatory and legal and C3b reputational and trust are components of C3. \\
I0 to I3 & Independence levels of evidence. \\
L0 to L3 & Control assurance levels: Present, Designed, Observed, Verified. \\
R0 to R3 & Action reversibility classes in an authority envelope. \\
T1 to T4 & Monitoring priority tiers; TR1 to TR6 are the tier rules. \\
EVR, SES, CSE & Evidence Validity Register; System Evaluation Specification (the per-system record); Continuous System Evaluation. \\
\bottomrule
\end{tabular}
\end{center}

\section{Correspondence with the Canonical Specifications}\label{app:sources}
This paper restates the methodology of the three version 2.1 documents (Public Framework, CSE Specification, NIST AI RMF crosswalk), in their corrected edition of 4 November 2026 whose original release was 29 October 2026, and does not alter its definitions, taxonomy, notation, caveats, decision rules or maturity labels. The table lists where each element of this paper is specified. Section numbers of the specifications are those of the corrected edition.
\begin{center}\small
\begin{tabular}{@{}L{5.0cm}L{4.2cm}L{5.8cm}@{}}
\toprule
Element (this paper) & Section here & Canonical source \\
\midrule
6+4+3 architecture, objectives, Register & \S\ref{sec:arch} & Public Framework: architecture, objectives, Consequence Register \\
Conditional identity, extensions, tail loss & \S\ref{sec:risk} & Public Framework: quantitative interpretation, materiality, stressed and tail loss \\
Statuses, independence, ladder, dispositions, rules & \S\ref{sec:evidence} & Public Framework: evidence architecture, dispositions and rules; CSE: control effectiveness \\
Lifecycle, EVR, invalidation & \S\ref{sec:lifecycle}, \S\ref{sec:evr} & CSE: lifecycle, Evidence Validity Register \\
Authority envelope, oversight & \S\ref{sec:controls} & CSE: agentic trajectory and authority, human oversight \\
Tiers, allocation, escalation, learning & \S\ref{sec:tiers} and \S\ref{sec:escalation} & CSE: monitoring, protocol, failure-to-learning loop \\
Policy-conditioned and avoided loss & \S\ref{sec:avoided} & CSE: quantitative foundation \\
Synthetic refund agent & \S\ref{sec:case} & Public Framework: worked application; CSE: illustration \\
Hypotheses H1 to H7 & \S\ref{sec:validation} & Validation Plan \\
\bottomrule
\end{tabular}
\end{center}
Differences of presentation (condensed tables, rephrased definitions) are listed in the accompanying source consistency report.

\section{Supplementary Specification Details}\label{app:details}
\paragraph{Further bounds.} Independent controls: if failures are independent, $\Prob(\text{all fail})=\prod_i p_i$; independence must be evidenced (separate code, data, vendors, failure causes). Combining intervals: $k$ intervals with confidence $c_i$ jointly cover with probability at least $1-\sum_i(1-c_i)$ (Bonferroni; conservative). Exceedance over exposure: $\Prob(\text{at least one})=1-(1-p)^n$ over $n$ independent exposures; $p=0.01$, $n=100$ gives $0.63$ (checked). Abstention: if evidence is too thin for a defensible estimate, record an explicit abstention with a reason, never zero; abstaining keeps the claim Unresolved.

\paragraph{Trajectory metrics.} Counts over scenarios, each reported with a Clopper--Pearson bound, none combined: unauthorized-action rate, unapproved consequential-action rate, irreversible action without human approval, side effect outside the envelope, delegation-authority violation rate and untrusted-memory-influenced consequential action (upper bounds); correct recovery after injected failure and required human handoff (lower bounds).

\paragraph{Human oversight measures.} Seeded-error detection (lower bound against a pre-set minimum); automation bias (upper bound on approval of known-wrong recommendations); false intervention (upper bound on rejection of correct recommendations, since a reviewer who blocks everything is not an effective control); dwell time and workload (a fast median with near-total approval indicates rubber-stamping); escalation speed against the harm-onset window; authority to act; independence. The result is Supported only if every criterion is met on its conservative bound, Refuted if an observed rate misses its threshold or a structural condition fails, and otherwise Unresolved.

\paragraph{Cadence defaults.} T1: daily signal check, weekly human review, monthly control retest, 90-day maximum authorization. T2: daily, monthly, quarterly, 180 days. T3: weekly, quarterly, semiannual, 365 days. T4: monthly, annual, annual, 365 days. Independent verification once per period for T1 and T2. When demand exceeds capacity, funding follows tier, then consequence class, then exposure, then identifier, with a reserved fraction for random audits; unfunded items are listed for the governance committee rather than moved to a lower tier.

\paragraph{Escalation defaults.} A completed harm event on a high-consequence claim: Stop pending review. Drill or retest failure of a credited control on a high-consequence claim: Escalate, and Stop if the control is the only barrier. Monitoring threshold exceeded (non-harm): Escalate to the risk owner within the tier's review period, and Restrict if the threshold concerns exposure. Evidence past its review date: Reassess; the claim may stay Supported only until the stated retest deadline and cannot support an expansion of authority. Evidence expired or invalidated: the claim reverts to Unresolved; no Deploy while tied to a high-consequence entry; Restrict only if a pathway-preventing control remains credited and evidenced. Repeated near misses on one pathway: Remediate. Reviewer or monitor capacity exceeded: Escalate.

\paragraph{Learning-loop outputs.} New evaluation scenarios, an updated failure taxonomy, revised control tests, changed consequence assumptions, monitoring improvements, corrective actions and new authorization evidence.

\paragraph{A hypothetical comparison (\synth{}).} All values here are invented for arithmetic. A fictional agent is compared in three configurations of one system, with the exposure unit a session and severity in arbitrary currency units: CFG-01 has no independent containment ($\Prob(F)=0.04$, $\Prob(H\cnd F)=0.80$, $\Ex[S]=60$, so $\Ex[L_F]=1.92$), CFG-02 adds an independent server-side limit check ($0.04$, $0.01$, $70$, so $0.028$), and CFG-03 adds reduced authority ($0.03$, $0.005$, $60$, so $0.009$). The failure rate is unchanged between CFG-01 and CFG-02 because the containment acts on $\Prob(H\cnd F)$, and severity is higher in CFG-02 because the harms that remain are the uncontained larger ones. The ordering holds only if each factor is estimated with evidence, and a small expected value does not make a prohibited or irreversible harm acceptable. The comparison says nothing about pathways $N$ and $M$.

\emph{Hypothetical arithmetic (\synth{}).} Take 100{,}000 sessions and one pathway with $\Prob(F)\in[0.001,0.003]$, $\Prob(H\cnd F)\in[0.2,0.4]$ and $\Ex[S]\in[50,100]$ currency units, so $X\in[0.01,0.12]$ per session. A hypothetical policy has $d\in[0.7,0.9]$, $t\in[0.6,0.8]$ and $c\in[0.6,0.8]$, so $dtc\in[0.252,0.576]$ and the residual factor lies in $[0.424,0.748]$. The coupled contrast is $[0.00252,0.06912]$ per session, or $[252,6912]$ over the horizon. Bounding the two losses separately instead gives $[-7976,11576]$, which includes zero and negative values and is wider because shared uncertainty is counted twice. These are interval computations, verified by the script of Appendix~\ref{app:repro}, not estimates. The contrast concerns expected loss only; a policy can lower expected loss and leave stressed loss unchanged when a common-cause burst exhausts response capacity, so the stress scenarios for high-consequence entries include the monitoring chain itself failing. A result may be called causal only for the tested conditions of a designed experiment (randomized exposure, a concurrent control arm or controlled failure injection); otherwise it is labeled by its evidence basis.
\begin{table}[h]
\centering\footnotesize
\caption{Four kinds of ongoing activity and what each establishes.}
\label{tab:activities}
\begin{tabular}{@{}L{3.2cm}L{5.6cm}L{6.0cm}@{}}
\toprule
Activity & Cadence & What it establishes \\
\midrule
Calendar-based review & At the end of a validity period or authorization expiry. & Whether evidence and authorization are still current. \\
Continuous monitoring (automated signal checks) & Per tier, from continuous or daily to monthly; only signals that must be fast need near-real-time evaluation. & Whether a pre-set threshold has been crossed. It does not show that controls work. \\
Scheduled challenge & Monthly to annual by tier: drills, forced-failure retests, seeded-failure detection tests. & Whether safeguards and monitors still work under realistic failure. \\
Event-triggered reassessment & On a change to a pinned component, an incident or near miss, a threshold breach, drift, an expansion of authority or expiry. & Whether specific evidence still holds after the event. \\
\bottomrule
\end{tabular}
\end{table}

\section{Hypotheses H1 to H7}\label{app:hyp}
\begin{table}[h]
\centering\footnotesize
\caption{Hypotheses H1 to H7 (summarized from the Validation Plan; paraphrased). None has been tested.}
\label{tab:hyp}
\begin{tabular}{@{}L{0.8cm}L{14.2cm}@{}}
\toprule
ID & Hypothesis \\
\midrule
H1 & Trained reviewers applying the SES rules agree on evidence status and the high-consequence classification of the same record more than chance. \\
H2 & Reconstructing documented incidents as SES records exposes the control, evidence or authorization gap later reported by investigators, more often than unstructured review. \\
H3 & Independent (I2 or I3) control challenge finds failures or gaps that I0 self-assessment of the same system misses. \\
H4 & Intervention-contrast estimates of control effectiveness predict containment observed later in monitored operation. \\
H5 & Teams can complete a record at an acceptable cost, and completing it changes at least one decision or safeguard. \\
H6 & The authorization rules block decisions that experts independently judge unsupported, without blocking many they judge supported. \\
H7 & Under a fixed monitoring budget, allocation by ordinal consequence-aware tiers avoids more expected harm than allocation by observed frequency or uniformly, where severity and frequency diverge and consequence is classified reliably; and it does no better, or worse, in specified other settings. \\
\bottomrule
\end{tabular}
\end{table}

\end{document}